\documentclass{article}

\usepackage[usenames]{color}
\definecolor{plum}  {rgb}{.4,0,.4}
\definecolor{forest}  {rgb}{0,.6,0}
\definecolor{midnight}  {rgb}{0,0,.8}
\definecolor{brick}  {rgb}{.8,0,0}
\usepackage[pdftex,plainpages=false,pdfpagelabels,colorlinks=true,urlcolor=midnight,linkcolor=brick,citecolor=midnight]{hyperref} 
\RequirePackage{hypernat}
\RequirePackage{epsfig, graphicx}
\RequirePackage{times}
\RequirePackage{latexsym}
\RequirePackage{psfrag}

\usepackage{amsthm,amsmath,amsfonts, amssymb}
\usepackage{fullpage}
\usepackage{epsf,epsfig,subfig}
\usepackage{fancyheadings}
\usepackage{graphics,graphicx}
\usepackage{enumerate}
\usepackage{epsfig}

\def\Poisson{\operatorname{Poisson}}

\def\cf{{\em cf.~}}
\def\ie{{\em i.e.,~}}
\def\eg{{\em e.g.,~}}

\theoremstyle{plain}

\newtheorem{theo}{Theorem}[section]

\newtheorem{lem}{Lemma}[section]
\newtheorem{prop}{Proposition}[section]
\newtheorem{cor}{Corollary}[section]

\theoremstyle{definition} 

\newtheorem{nota}{Notation}[section]
\newtheorem{de}{Definition}[section]
\newtheorem{exa}{Example}[section]
\newtheorem{as}{Assumption}[section]
\newtheorem{alg}{Algorithm}[section]

\newcommand{\btheo}{\begin{theo}}
\newcommand{\bde}{\begin{de}}
\newcommand{\ble}{\begin{lem}}
\newcommand{\bpr}{\begin{prop}}
\newcommand{\bno}{\begin{nota}}
\newcommand{\bex}{\begin{exa}}
\newcommand{\bcor}{\begin{cor}}
\newcommand{\spro}{\begin{proof}}
\newcommand{\bas}{\begin{as}}
\newcommand{\balg}{\begin{alg}}

\newcommand{\etheo}{\end{theo}}
\newcommand{\ede}{\end{de}}
\newcommand{\ele}{\end{lem}}
\newcommand{\epr}{\end{prop}}
\newcommand{\eno}{\end{nota}}
\newcommand{\eex}{\end{exa}}
\newcommand{\ecor}{\end{cor}}
\newcommand{\fpro}{\end{proof}}
\newcommand{\eas}{\end{as}}
\newcommand{\ealg}{\end{alg}}

\theoremstyle{plain}

\newtheorem{theos}{Theorem}
\newtheorem{props}{Proposition}
\newtheorem{lems}{Lemma}
\newtheorem{cors}{Corollary}

\theoremstyle{definition}
\newtheorem{exas}{Example}
\newtheorem{algs}{Algorithm}
\newtheorem{asss}{Asumption}
\newtheorem{defns}{Definition}

\newcommand{\btheos}{\begin{theos}}
\newcommand{\etheos}{\end{theos}}
\newcommand{\bprops}{\begin{props}}
\newcommand{\eprops}{\end{props}}
\newcommand{\bdes}{\begin{defns}}
\newcommand{\edes}{\end{defns}}
\newcommand{\blems}{\begin{lems}}
\newcommand{\elems}{\end{lems}}
\newcommand{\bcors}{\begin{cors}}
\newcommand{\ecors}{\end{cors}}
\newcommand{\bexs}{\begin{exas}}
\newcommand{\eexs}{\end{exas}}
\newcommand{\balgs}{\begin{algs}}
\newcommand{\ealgs}{\end{algs}}
\newcommand{\bass}{\begin{asss}}
\newcommand{\eass}{\end{asss}}

\newcommand{\Prob}{\mathbb{P}}

\DeclareMathOperator*{\argmin}{arg\,min}
\newcommand{\T}{\top}

\title{Inference of High-dimensional Autoregressive Generalized Linear Models}

\author{Eric~C.~Hall\thanks{E. C. Hall is with the Wisconsin Institute of Discovery, University of Wisconsin-Madison, Madison, WI, 53706, USA. e-mail:
    echall@wisc.edu},
 Garvesh Raskutti\thanks{G. Raskutti is with the Department of Statistics, University of Wisconsin-Madison, Madison, WI, 53706, USA. e-mail: raskutti@stats.wisc.edu},  
  and~Rebecca~M.~Willett\thanks{R. M. Willett is with the Department of
    Electrical and Computer Engineering, University of
    Wisconsin-Madison, Madison, WI 53706, USA. e-mail:
    willett@discovery.wisc.edu.
    We gratefully acknowledge the support of the awards NSF CCF-1418976,
NIH 1 U54 AI117924-01,
14-AFOSR-1103, and NSF DMS-1407028.}}

\begin{document}
\maketitle

\begin{abstract}
  Vector autoregressive models characterize a variety of time series
  in which linear combinations of current and past observations can be
  used to accurately predict future observations. For instance, each
  element of an observation vector could correspond to a different
  node in a network, and the parameters of an autoregressive model
  would correspond to the impact of the network structure on the time
  series evolution.  Often these models are used successfully in
  practice to learn the structure of social, epidemiological,
  financial, or biological neural networks.  However, little is known
  about statistical guarantees on estimates of such models in
  non-Gaussian settings.  This paper addresses the inference of the
  autoregressive parameters and associated network structure within a
  generalized linear model framework that includes Poisson and
  Bernoulli autoregressive processes.  At the heart of this analysis
  is a sparsity-regularized maximum likelihood estimator. While
  sparsity-regularization is well-studied in the statistics and
  machine learning communities, those analysis methods cannot be
  applied to autoregressive generalized linear models because of the
  correlations and potential heteroscedasticity inherent in the
  observations.  Sample complexity bounds are derived using a
  combination of martingale concentration inequalities and modern
  empirical process techniques for dependent random variables. These
  bounds, which are supported by several simulation studies,
  characterize the impact of various network parameters on estimator
  performance.
\end{abstract}
\section{Autoregressive Processes in High Dimensions} 

Imagine recording the times at which each neuron in a biological
neural network fires or ``spikes''. 
Neuron spikes can
trigger or inhibit spikes in neighboring neurons, and understanding
excitation and inhibition among neurons provides key insight into the
structure and operation of the underlying neural network \cite{brown2004multiple,colemanConvexPoint,SmithBrownStateSpace,HinneHeskes2012,
  DingSchroeder2011,spikesPillow,spikesMasud}. A central question in
the design of this experiment is {\em ``for how long must I collect
  data before I can be confident that my inference of the network is
  accurate?''} Clearly the answer to this question will depend not
only on the number of neurons being recorded, but also on what we may
assume {\em a priori} about the network. Unfortunately, existing
statistical and machine learning theory give little insight into this
problem. 

Neural spike recordings are just one example of a non-Gaussian,
high-dimensional autoregressive processes, where the autoregressive
parameters correspond to the structure of the underlying network.
This paper examines a broad class of such processes, in which each
observation vector is modeled using an exponential family
distribution. In general, autoregressive models are a widely-used
mechanism for studying time series in which each observation depends
on the past sequence of observations. Inferring these dependencies is
a key challenge in many settings, including finance, neuroscience,
epidemiology, and sociology. A precise understanding of these
dependencies facilitates more accurate predictions and interpretable
models of the forces that determine the distribution of each new
observation.

Much of the autoregressive modeling literature focuses on Gaussian
noise and perturbation models, but in many settings Gaussian noise
fails to capture the data at hand. This challenge arises, for
instance, when observations correspond to count data -- \eg when we
collect data by counting individual events such as neurons spiking.
Another example arises in epidemiology, where a common model involves
infection traveling stochastically from one node in a network to
another based on the underlying network structure in a process known
as an ``epidemic cascade''
\cite{netrapalli2012learning,altarelli2014patient,kempe2003maximizing,kuperman2001small}.
These models are used to infer network structure based on the
observations of infection time, which is closely related to the
Bernoulli autoregressive model studied in this paper.
Further examples arise in a variety of applications, including
vehicular traffic analysis
\cite{johansson1996speed,matteson2011forecasting}, finance
\cite{rydberg1999modelling,ait2010modeling,chavez2012high,cameron2013regression},
social network analysis
\cite{raginsky_OCP,silva:pami,BertozziHawkes,HellerHawkes,zhouZhaSongHawkes},
biological neural networks
\cite{brown2004multiple,colemanConvexPoint,SmithBrownStateSpace,HinneHeskes2012,
  DingSchroeder2011,spikesPillow,spikesMasud}, power systems
analysis \cite{huang2003short}, and seismology
\cite{hawkesEarthquake,ogata1999seismicity}.  

Because of their
prevalence across application domains, time series count data 
(\cf~\cite{brannas1994time,macdonald1997hidden,zeger1988regression,jorgensen1999state,fahrmeir2013multivariate})
and other non-Gaussian autoregressive processes (\cf~\cite{grunwald2000theory,benjamin2003generalized,gourieroux2005autoregressive})
have
been studied for decades.
Although a substantial fraction of the this literature is focused on univariate
time series, this paper focuses on multivariate settings, particularly
where the vector observed at each time is high-dimensional relative to
the duration of the time series. In the above examples, the dimension of
the each observation vector would be the number of neurons in a
neural network, the number of people in a social network, or the
number of interacting financial instruments.

In this paper, we conduct a detailed investigation of a particular
family of time series that we call the {\em vector generalized linear
  autoregressive} (GLAR) model. In addition, we examine our results
for two members of this family: the Bernoulli autoregressive and the
log-linear Poisson autoregressive (PAR) model.  The PAR model has been
explicitly studied in
\cite{fokianos2009poisson,zhu2011estimation,fokianos2011log} and is
closely related to the continuous-time Hawkes point process model
\cite{hawkes1,hawkes2,PointProcesses,hansenReynaud2015, bacry2015} and the
discrete-time INGARCH model
\cite{heinen2003modelling,zhu2011negative,zhu2012modeling,zhu2012modeling_b}. However,
that literature does not contain the sample complexity results
presented here. The INGARCH literature is focused on
  low-dimensional settings, typically univariate, whereas we are
  focused on the high-dimensional setting where the number of nodes or
  channels is high relative to the number of
  observations. Additionally, existing sample complexity bounds for Hawkes
  processes \cite{hansenReynaud2015} focus on a linear (as opposed to
  log-linear) model with samples collected after reaching the
  stationary distribution. The log-linear model is largely used in
  practice both for numerical reasons and modeling efficacy for real
  world data. We note that linear models can predict inadmissible
  negative event rates, whereas the log-linear model enforces the
  feasibility of the predicted model. The log-linear and linear models
  exhibit very different behaviors in their properties and stationary
  distributions, making this work a significant step forward from the
  analysis of linear models. The extension of these prior investigations to the high-dimensional, non-stationary setting is non-trivial and requires the development of new theory and methods.

This paper focuses on estimating the parameters of a vector GLAR model
from a time series of observations. We adopt a regularized likelihood
estimation approach that extends and generalizes our previous work on Poisson inverse
problems
(\cf~\cite{willett:density,dense_pcs,expander_PCS,JiangWillettRaskutti15}). While
similar algorithms have been proposed in the above-mentioned 
literature, little is known about their {\em sample complexity} or
{\em how inference accuracy scales with the key parameters such as the
  size of the network or number of entities observed, the time spent
  collecting observations, and the density of edges within the network
  or dependencies among entities.}  

There has been a large body of work providing theoretical results for
certain high-dimensional models under low-dimensional structural
constraints (see \eg \cite{geer08, JiangWillettRaskutti15, KolYua08,
  Meier09, Neg10, RasWaiYu11, RasWaiYu12, Zhao06, bacry2015}).  The majority of
prior work has focused on the setting where samples are independent
and/or follow a Gaussian distribution.  In the GLAR setting, however,
non-Gaussianity and temporal dependence among observations can make
such analyses particularly challenging and beyond the scope of much
current research in high-dimensional statistical inference (see
\cite{BvdG2011} for an overview).

Perhaps the most closely related prior work to our setting in the
high-dimensional setting is ~\cite{BasuMichail15}. In
~\cite{BasuMichail15}, several performance guarantees are provided for
different linear Gaussian problems with dependent samples including
the Gaussian autoregressive model. Since~\cite{BasuMichail15} deals
exclusively with linear Gaussian models, they exploit many properties
of linear systems and Gaussian random variables that cannot be applied
to non-Gaussian and non-linear autoregressive models.  In particular,
compared to standard autoregressive processes with Gaussian noise, in
the GLAR setting the conditional variance of each observation is
dependent on previous data instead of being a constant equal to the
noise variance.  Works such
as~\cite{JiangWillettRaskutti15,geer08,jiang2015data} provide results
for non-Gaussian models but still rely on independent observations.
Weighted LASSO estimators for Hawkes processes address some of these
challenges in a continuous-time setting \cite{hansenReynaud2015}.

To see why GLAR analysis can be challenging, consider momentarily a
LASSO estimator of the autoregressive parameters. In the classical
LASSO setting, the accuracy of the estimate depends on characteristics
of the Gram matrix associated with the design or sensing matrix. This
matrix may be stochastic, but it is usually considered independent of
the observations and performance guarantees for the estimator depend
on the assumption that the matrix obeys certain properties (\eg the
restricted eigenvalue condition \cite{BiRiTsy08}). In our setting,
however, the ``design'' matrix is a function of the observed data,
which in turn depends on the true underlying network or autoregressive
model parameters.  Thus a key challenge in the analysis of a
LASSO-like estimator in the GLAR setting involves showing that the
data- and network-dependent Gram matrix exhibits properties that
ensure reliable estimates.

In this paper, we develop performance guarantees for the vector GLAR
model that provide sample complexity guarantees in the
high-dimensional setting under low-dimensional structural assumptions
such as sparsity of the underlying autoregressive parameters.
In particular, our main contributions are the following:
\sloppypar \begin{itemize}
  \item Formulation of a maximum penalized likelihood estimator for
  vector GLAR models in high-dimensional settings with sparse
  structure.
\item Mean-squared-error bounds on the proposed estimator as a
  function of the problem dimension, sparsity, and the number of
  observations in time for general GLAR models.
\item Application of our general result to obtain sample complexity bounds
for Bernoulli and Poisson GLAR models.  
\item Analysis techniques that simultaneously leverage martingale concentration
  inequalities, empirical risk minimization analysis, and covering
  arguments for high-dimensional linear regression. 
\end{itemize}
This problem is substantially harder than the Gaussian case from
  a technical perspective because we can not exploit linearity and
  spectral properties of linear Gaussian time-series. In our case we
  have signal-dependent noise, and we can not exploit the same
  spectral properties. Additionally, with non-Gaussian noise, we are
  not guaranteed strong convexity of the objective function in the
  entire domain of possible solutions, and so extra care must be taken
  to define regions of strong convexity. Thus we have to develop new
  theoretical techniques, using new concentration bounds and a more
  refined analysis. The remainder of the paper is structured as
follows: Section \ref{ARProbForm} introduces the generalized linear
autoregressive model and Section \ref{ARMSEBounds} presents the novel
risk bounds associated with the RMLE of the process. We then use our
theory to examine two special cases (the Poisson and Bernoulli models)
in Sections \ref{subBern} and \ref{subPois}, respectively. The main
proofs are provided in Section~\ref{sec:proofs}, while supplementary
lemmas are deferred to the appendix. Finally,
Section~\ref{sec:discussion} contains a discussion of our results,
their implications in different settings, and potential avenues for
future work.

\section{Problem Formulation}
\label{ARProbForm}

In this paper we consider the generalized linear autoregressive model:
\begin{equation}
\label{EqnModel}
X_{t+1,m}| X_t \sim p(\nu_m + a_m^{*\top} X_t),
\end{equation}
where $X_{t+1,m}$ is the $m^{th}$ observation of $X_{t+1}$, $(X_t)_{t=0}^{\infty}$ are $M$-variate vectors and
$a^*\in [a_{\min}, a_{\max}]^{M}$ is an unknown parameter
vector, $\nu \in [\nu_{\min},\nu_{\max}]^M$ is a known, constant
offset parameter,
  and $p$ is an exponential family probability
  distribution. Specifically, $X \sim p(\theta)$ means that the
  distribution of the scalar $X$ is associated with the density
  $p(x|\theta) = h(x) \exp[\phi(x)\theta-Z(\theta)]$, where $Z(\theta)$ is
  the so-called {\em log partition function}, $\phi(x)$ is the sufficient statistic
  of the data, and $h(x)$ is the base measure of the distribution. Distributions that fit such assumptions include the Poisson,
Bernoulli, binomial, negative binomial and exponential. According to this model, conditioned on the previous data,
  the elements of $X_t$ are independent of one another and each have a scalar natural parameter. The input of the
function $p$ in \eqref{EqnModel} is the natural parameter for the
distribution, \ie ~$\nu + a_m^{*\top}X_t$ is the natural parameter of the
conditional distribution at time $t+1$ for observation $m$.  A similar, but low-dimensional, model appears in
\cite{fokianos2011log}, but that work focuses on maximum likelihood
and weighted least squares estimators in univariate settings that are
known to perform poorly in high-dimensional settings (as is our
focus). For these distributions it is
straightforward to show when they have
strongly convex
log-partition functions, which will be crucial to our
analysis. 
Note that this distribution has $\mathbb{E}[\phi(X_{t+1,m})|X_t] = Z'(\nu + a_m^{*\top}X_t)$ and
  $ {\rm{Var}} (\phi(X_{t+1,m})|X_t) = Z''(\nu + a_m^{*\top}X_t)$, the first and second derivatives of the 
  log-partition function, respectively.  Compared to standard
autoregressive processes with Gaussian noise, the conditional variance
is now dependent on previous data instead of being a constant equal to
the noise variance.

We can state the conditional distribution explicitly as:
\begin{equation*}
\mathbb{P}(X_{t+1} | X_t ) = \prod_{m=1}^M h(X_{t+1,m}) \exp\left(\phi(X_{t+1,m}) (\nu_m + a_m^{*\top}X_t) - Z(\nu_m + a_m^{*\top}X_t) \right),
\end{equation*}
where $h$ is the base-measure of the distribution $p$. Using this equation and observations, we can find an estimate for the network $A^*$ which is constructed row-wise by $a_m^*$. ($i.e.$ $a_m^{*\top}$ is the $m^{th}$ row of $A^*$).

In general, we observe $T$ samples $(X_t)_{t=0}^T$ and our goal is to
infer the matrix $A^*$. In the setting where $M$ is large, we need to
impose structural assumptions on $A^*$ in order to have strong
performance guarantees. Let 
$$\mathcal{S} :=
\{(\ell,m) \in \{1,\ldots,M\}^2 : A^*_{\ell,m} \neq 0 \}.$$
In this paper we assume that the matrix
$A^*$ is $s$-sparse, meaning that $A^*$ belongs to the following
class:
\begin{equation*}
\mathcal{A}_s = \left\{A \in [a_{\min},a_{\max}]^{M \times M}\;|\; \|A\|_0 \leq s \right\}.
\end{equation*}
where $\|A\|_0 := \sum_{\ell = 1}^M \sum_{m=1}^M
\mathbf{1}(|A_{\ell,m}| \neq 0)$ and $\mathbf{1}(\cdot)$ is the
indicator function. That is, we assume $|\mathcal{S}| = s$.
Furthermore, we define $$\rho_m \triangleq \displaystyle \|a_m^*\|_0
\qquad \mbox{ and } \qquad \rho \triangleq \displaystyle\max_m
\rho_m,$$ so $\rho$ is the maximum number of non-zero
elements in a row of $A^*$. 

We might like to estimate $A^*$ via a constrained maximum likelihood
estimator by solving the following optimization problem:
\begin{equation}
\argmin_{A \in \mathcal{A}_s}\frac{1}{T}\sum_{t=0}^{T-1} \sum_{m=1}^M{\biggr( Z(\nu_m + a_m^{\T}X_t) - a_m^{\T}X_t \phi(X_{t+1,m}}) \biggr) 
\end{equation}	
or its Lagrangian form
\begin{equation}
\argmin_{A \in [a_{\min}, a_{\max}]^{M \times M}}\frac{1}{T}\sum_{t=0}^{T-1} \sum_{m=1}^M{\biggr( Z(\nu_m + a_m^{\T}X_t) - a_m^{\T}X_t \phi(X_{t+1,m}}) \biggr) + \lambda \|A\|_0.
\end{equation}	
 However, these are difficult optimization problems due to the
non-convexity of the $\ell_0$ norm. Therefore, we instead find an
estimator using the element-wise $\ell_1$ regularizer,
the convex relaxation of the $\ell_0$ function, along with the
negative log-likelihood to create the following estimator:
\begin{equation}
\label{EqnLogLike}
\widehat{A} = \argmin_{A \in [a_{\min}, a_{\max}]^{M \times M}}\frac{1}{T}\sum_{t=0}^{T-1} \sum_{m=1}^M{\biggr( Z(\nu_m + a_m^{\T}X_t) - a_m^{\T}X_t \phi(X_{t+1,m}}) \biggr) + \lambda \|A\|_{1,1},
\end{equation}	
where $\|\cdot\|_1$ is the $\ell_1$ norm and $\|A\|_{1,1}
=\sum_{m=1}^M \|a_{m}\|_1$. The above is the regularized maximum
likelihood estimator (RMLE) for the problem, which attempts to find an
estimate of $A^*$ which both fits the empirical distribution of the
data while also having many zero-valued elements. Notice that we assume the elements of $A^*$ are bounded and we use these bounds in the estimator definition. One reason for this is that bounds on the elements of $A^*$ can enforce stability. If the elements of $A^*$ are allowed to be arbitrarily large, the system may become unstable and therefore impossible to make proper estimates. Knowing loose bounds facilitates our analysis but in practice does not appear to be necessary. In the experiment section we discuss choosing these bounds in the estimation process.

We note that while we assume that $\nu$ is a known constant vector, if we assume there is some
unknown constant offset that we would like to estimate, we can fold it into the estimation of $A$. For instance, consider appending $\nu$ as an extra column of the matrix $A^*$, and appending a 1 to the end of each observation $X_t$. Then for indices $1,\ldots,M$ the observation model becomes $X_{t+1,m}|X_t \sim p(a_m^{*\top}X_t)$ where $a_m^*$ and $X_t$ are the appended versions. We can then find the RMLE of this distribution to find both $\widehat{A}$ and $\widehat{\nu}$, but for clarity of exposition we assume a known $\nu$.

Estimating the network parameters in the autoregressive setting with Gaussian observations can be formulated as a sparse inverse problem with connections to the well-known LASSO
estimator.  Consider the problem of estimating the $a^*_m$. Define
\begin{equation*}
y_{m} = \begin{bmatrix}
X_{2,m}\\ X_{3,m} \\ \vdots \\ X_{T,m} \end{bmatrix}
\qquad \mbox{ and } \quad
\bf{X} = \begin{bmatrix}
X_{1,1} & X_{1,2} & \cdots & X_{1,M}\\
X_{2,1} & X_{2,2} & \cdots & X_{2,M}\\
\vdots & \vdots & \ddots \vdots \\
X_{T-1,1} & X_{T-1,2} & \cdots & X_{T-1,M}
\end{bmatrix},
\end{equation*}
where $y_m$ is the time series of observed counts associated with the
$m^{\rm th}$ node and ${\bf{X}}$ is a matrix of the observed counts
associated with all nodes. Then 
$y_m = \bf{X} a^*_m + \epsilon_m,$
where $\epsilon_m := y_m - \bf{X} a^*_m$ is noise, and we could consider the
LASSO estimator for each $m$:
\begin{equation*}
\hat a_m = \argmin_a \|y_m - \bf{X} a\|_2^2 + \lambda \|a\|_1.
\end{equation*}
However, there are two key challenges associated with the LASSO estimator in this context:
 (a) the squared residual term does not account for
the non-Gaussian statistics of the observations and (b) the ``design
matrix'' is data-dependent and hence a function of the unknown
underlying network. In classical LASSO analyses, performance bounds depend on the
design matrix satisfying the {\em restricted eigenvalue condition} or
{\em restricted isometry property} or some related condition; it is
relatively straightforward to ensure such a condition is satisfied
when the design matrix is independent of the data, but much more
challenging in the current context. As a result, despite the fact that
we face a sparse inverse problem, the existing LASSO literature does
not address the subject of this proposal.

\section{Main Results}
\label{ARMSEBounds}
In this section, we turn our attention to deriving bounds for
$\|\widehat{A} - A^*\|_F^2$, the difference in Frobenius norm between the regularized
maximum likelihood estimator, $\widehat{A}$, and the true generating
network, $A^*,$ under the assumption that the true network is
sparse. We assume that $A^* \in \mathcal{A}_s$. Recall 
$\rho \triangleq \displaystyle\max_m \|a_m^*\|_0$ is the maximum number of non-zero
elements in a row of $A^*$. First we state assumptions on the GLAR process which are sufficient conditions to ensure the RMLE admits small errors. 

\bass
\label{AsLearnable}
We assume that for any realization of the process defined by Equation \ref{EqnModel} there
exists a subset of observations $\{X_{\mathcal{T}_t}\}_{t=1}^{|\mathcal{T}|}$ for $\mathcal{T} \subseteq \{0,1,\ldots,T-1\}$ that satisfies the conditions:
\begin{enumerate}
\item There exists a constant $U$ such that $U \geq \max_{t \in \mathcal{T}} \|X_t\|_\infty$ where $U$ is independent of $T$. 
\item$Z(\cdot)$ is $\sigma$-strongly convex on a domain determined by $U$:
$$ Z(x) \geq Z(y) + Z'(y)(x-y) + \frac{\sigma}{2}\|x-y\|_2^2$$
for all $x,y \in [-\tilde{\nu} - 9\rho \tilde{a}U, \tilde{\nu} + 9\rho
\tilde{a} U ]$ where $\tilde{\nu} \triangleq
\max(|\nu_{\min}|,|\nu_{\max}|),$ and $ \tilde{a} \triangleq
\max(|a_{\min}|,|a_{\max}|)$, where  $\sigma$ is independent of
$T$. 
\item The smallest eigenvalue of 
$\Gamma_t \triangleq \mathbb{E}[X_{\mathcal{T}_t}
X_{\mathcal{T}_t}^\top |X_{\mathcal{T}_{t-1}}]$ is lower bounded by $\omega > 0$, which is independent of $T$. 
\end{enumerate}
We define the constant $\xi$ as a constant such that $\xi \leq \triangleq |\mathcal{T}|/T$, which will be determined
in part by the constant $U$, and can be set such that $\xi$ is very
close to 1.  \eass
For $\xi \approx 1$, Assumption~\ref{AsLearnable} means most of the
observed data is bounded independent of $T$. 
The assumption allows us to analyze time series in which the maximum
of a series of iid random variables can grow with $T$, but any
percentile is bounded by a constant. Our analysis will then be
conducted on the bounded series $\{X_{\mathcal{T}_t}\}_{t=1}^{|\mathcal{T}|}$.
The
assumptions are proven to be true with high probability for the
Bernoulli and Poisson cases in Sections \ref{subBern} and
\ref{subPois}, respectively, and the corresponding values of $U$,
$\sigma$, $\xi$, and $\omega$ are computed explicitly.

\btheos
\label{ThmMain}
Assume $\lambda \geq \max_{1\leq m \leq M} \frac{2}{T}\left\|\sum_{t=0}^{T-1} {\Big(\phi(X_{t+1,m}) -
  \mathbb{E}[\phi(X_{t+1,m})|X_t]\Big) X_t}\right\|_\infty$, and let $\widehat{A}$ be
the RMLE for a process which obeys the conditions of Assumption \ref{AsLearnable}. For any row of the estimator and for any $\delta \in (0,1)$, with probability at least $1- \delta$,
\begin{equation*}
\|\widehat{a}_m - a_m^*\|_2^2 \leq \frac{144}{\xi^2\sigma^2\omega^2} \rho_m \lambda^2
\end{equation*}
for
$T \geq \frac{c\rho_m^2}{\omega^2} \left(\frac{\rho_m \log(2M)}{\omega^2} +
  \log(1/\delta) \right) $ where
$c$ is independent of $M,T,\rho$ and $s$.
Furthermore,
\begin{equation*}
\|\widehat{A} - A^*\|_F^2 \leq \frac{144}{\xi^2 \sigma^2 \omega^2} s \lambda^2
\end{equation*}
with probability greater than $1-\delta$ for $T \geq \frac{c \rho^2}{\omega^2} \left( \left(\frac{\rho }{\omega^2} + 1\right) \log(2M)+ \log(1/\delta) \right)$.
\etheos 

To apply Theorem~\ref{ThmMain} to specific GLAR models, we need to
provide bounds on $\lambda$, as well as $\sigma$, $\omega, U$ and $\xi$
for Assumption \ref{AsLearnable}. We do this in the next section for Bernoulli and Poisson GLAR models.

We can compare the results of Theorem~\ref{ThmMain} to the
  related results of \cite{BasuMichail15}. In that work they arrive at
  rates for the Gaussian autoregressive process that are equivalent
  with respect to the sparsity parameter, number of observations and
  regularization parameter. However, we incur slightly different
  dependencies on $\xi, \sigma$ and $\omega$. These are due mainly to
  the fact that our bounds hold for a wide family of distributions and
  not just the Gaussian case, which has nice properties related to
  restricted strong convexity and specialized concentration
  inequalities. Additionally, the way $\lambda$ is defined is very
  similar, but bounding $\lambda$ for a non-Gaussian distribution will
  result in extra log factors. It is an open question whether this
  bound is rate optimal in the general setting.

\subsection{Example 1: Bernoulli Distribution}
\label{subBern}
For the Bernoulli distribution we have the following autoregressive model:
\begin{align}
X_{t+1,m}|X_t \sim {\rm{Bernoulli}}\left(\frac{1}{1+\exp(-\nu - a_m^{*\top}X_t)}\right). \label{eq:Bern}
\end{align}
The first observation about this model is that the sufficient statistic $\phi(x) = x$ and the log-partition
function $Z(\theta) = \log(1+\exp(\theta))$, which is strongly convex
when the absolute value of $\theta$ is bounded. One advantage of this
model is that the observations are inherently bounded due to the
nature of the Bernoulli distribution, so $\mathcal{T} = [0,1,\ldots, T-1]$ and
 $\xi = 1$. Using this observation we derive the strong
convexity parameter of $Z$ on the bounded range, thus
$\sigma = (3 + \exp(\tilde{\nu} + 9 \rho \tilde{a}))^{-1}$. 

To derive rates from Theorem~\ref{ThmMain}, we must prove that Assumption \ref{AsLearnable} holds; this is shown with high-probability by Theorem~\ref{ThmBern}.
\btheos
\label{ThmBern}
For a sequence $X_t$ generated from the Bernoulli autoregressive process with the matrix $A^*$ with and the vector $\nu$, we have the following properties:
\begin{enumerate}
\item The smallest eigenvalue of the matrix $\Gamma_t  =
  \mathbb{E}[X_{t}X_{t}^\T|X_{{t-1}}]$ is lower bounded by
  $\omega = (3 + \exp(\tilde{\nu} + \rho \tilde{a}))^{-1}$.
\item
Assuming $1 \leq t \leq T$ and that $T \geq 2$ and $\log(MT) \geq 1$, then
$$\max_{1 \leq i,j \leq M} \frac{1}{T}\left|\sum_{t=0}^{T-1} X_{t-1,i} (X_{t,j} - \mathbb{E}[X_{t,j} | X_{t-1}])\right| \leq \frac{3 \log(MT)}{\sqrt{T}}$$
with probability at least at least $1-\frac{1}{MT}$.
\end{enumerate}
\etheos

Using these results we get the final sample error bounds for the Bernoulli autoregressive process.
\bcors
\label{CorrBern}
The RMLE for the Bernoulli autoregressive process defined by Equation \ref{eq:Bern}, and setting $\lambda =  \frac{6\log(MT)}{\sqrt{T}}$ has error bounded by
$$ \|A^* - \widehat{A} \|_F^2 \leq C \left(3 + e^{\tilde{\nu} + 9 \rho \tilde{a}} \right)^4 \frac{s \log^2(MT)}{\xi^2 T}$$
with probability at least $1-\delta$ for $T \geq \max
\left(\frac{2}{\delta M}, \frac{c \rho^2}{\omega^2} \left(\left(1 + \frac{\rho
     }{\omega^2}\right) \log(2M) + \log(2/\delta) \right) \right)$ for constants $C, c > 0$ which are independent of $M, T, s$ and $\rho$.
\ecors
The lower bound on the number of observations $T$ comes from needing to satisfy the conditions of both parts of Theorems \ref{ThmMain} and \ref{ThmBern}. In order to get this statement we use a union bound over the high probability statements of Theorem \ref{ThmMain} described in \eqref{eq:ProbBound} and Theorem \ref{ThmBern} which holds with probability greater than $1 - \frac{1}{MT}$.

\subsection{Example 2: Poisson Distribution}
\label{subPois}
In this section, we derive the relevant values to get error bounds for the vector autoregressive Poisson distribution. Under this model we have
\begin{align*}
X_{t+1,m}|X_t \sim {\rm{Poiss}}(\exp(\nu + a_m^{*\top}X_t)).
\end{align*}
We assume that $a_{\max} = 0$ for stability purposes, thus we are only modeling inhibitory relationships in the network. Deriving the sufficient statistic and log-partition function yields $\phi(x) = x$ and $Z(\theta) = \exp(\theta)$. The next important values are the bounds on the magnitude of the observations, which will both ensure the strong convexity of $Z$ and the stability of the process.

\blems
\label{LemPoissBound}
For the Poisson autoregressive process generated with $A^* \in
[a_{\min},0]^{M \times M}$ and constant vector $\nu \in [\nu_{\min},
\nu_{\max}]$: 
\begin{enumerate}
\item
If $\log MT \geq 1$, there exists constants $C$ and $c$ which depend on the value $\nu_{\max}$, but are independent of $T,M,s$ and $\rho$ such that $0 \leq X_{t,m} \leq C \log (MT)$ with probability at least $1 - e^{-c \log(MT)}$ for all $1\leq t \leq T$ and $1 \leq m \leq M$.
\item
For any $\alpha \in (0,1)$ such that $\alpha MT$ is an integer, there
exist constants $U$ and $c$ which depend on the values of $\nu_{\max}$
and $\alpha$, but independent of $T,M,s$ and $\rho$, such that with probability at least $1 - e^{-c MT}$, $0 \leq X_{t,m} \leq U$ for at least $\alpha M T$ of the indices. We define $\mathcal{T}$ to be these $\alpha M T$ indices.
\end{enumerate}
\elems

As a consequence of Lemma \ref{LemPoissBound}, we have $\|X_t \|_\infty \leq U$ for at 
least $\xi T$ values of $t \in \{1,2,\ldots,T\}$ where $\xi = 1 - (1-\alpha)M$. We additionally assume that $U$ is
large enough such that $\alpha > \frac{M-1}{M}$ and therefore $\xi \in
(0,1)$. 

Using this Lemma,we prove that Assumption \ref{AsLearnable} holds with high-probability, by deriving the strong convexity parameter of $Z$ and a lower bound on the smallest eigenvalue of $\Gamma_t$. In the Poisson case, $Z(\cdot) = \exp(\cdot)$ and therefore the strong convexity parameter, $\sigma = \exp(-\tilde{\nu} +9 \rho a_{\min} U)$.
\btheos
\label{ThmPoiss}
For a sequence $X_t$ generated from the Poisson autoregressive process with the matrix $A^*$, with all non-positive elements, and the vector $\nu$, we have the following properties
\begin{enumerate}
\item
The smallest eigenvalue of the matrix $\Gamma_t  =
\mathbb{E}[X_{\mathcal{T}_t}X_{\mathcal{T}_t}^\T|X_{\mathcal{T}_{t-1}}]$, for consecutive indices
$\mathcal{T}_t$ and $\mathcal{T}_{t-1}$ in $\mathcal{T}$ as defined in Assumption \ref{AsLearnable}, is lower bounded by $\frac{4\xi}{5}\exp(\nu_{\min} + \rho a_{\min} U)$.
\item
Assuming $X_{t,m} \leq C \log(MT)$ for all $1\leq m \leq M$ and $1 \leq t \leq T$ and that $T \geq 2$ and $\log(MT) \geq 1$, then
$$\max_{1 \leq i,j \leq M} \frac{1}{T}\left|\sum_{t=0}^{T-1} X_{t-1,i} (X_{t,j} - \mathbb{E}[X_{t,j} | X_{t-1}])\right| \leq 4 C^2 e^{\nu_{\max}} \frac{\log^3(MT)}{\sqrt{T}} $$
with probability at least at least $1-\exp(-c \log(MT))$ for some $c>0$ independent of $\rho,s,M$ and $T$.
\end{enumerate}
\etheos
Using Theorem \ref{ThmPoiss}, we can find the error bounds for the PAR process by using the result of Theorem \ref{ThmMain}.

\bcors
\label{CorrPoiss}
Using the results of Theorem \ref{ThmMain} and using the Poisson autoregressive model with $A^*$ with all non-positive values, the RMLE admits the overall error rate of 
$$ \|\hat{A} - A^*\|_F^2 \leq C \exp(20 |a_{\min}| U\rho)\frac{s \log^6(MT)}{\xi^3 T}$$
with probability at least $1 - \delta$ for $T \geq \max\left(\left(\frac{4}{\delta M}\right)^c, \frac{c \rho^2}{\omega^2} \left(\left(\frac{\rho}{\omega^2} + 1\right)\log(2M) + \log(4/\delta) \right) \right)$ for constants $C, c > 0$ which are independent of $M, T, s$ and $\rho$

\ecors
Again, the lower bound on the number of observations comes from combining the high probability statements of each of the constituent parts of the corollary in the same way as was done in the Bernoulli case. In this case all of Theorem \ref{ThmMain}, both parts of Lemma \ref{LemPoissBound} and Theorem \ref{ThmPoiss} need to hold.

\subsection{Experimental Results}
\label{ARExpResults}
\begin{figure}[t!]
    \centering
    \subfloat[MSE vs. $T$]{\includegraphics[height=1.5in]{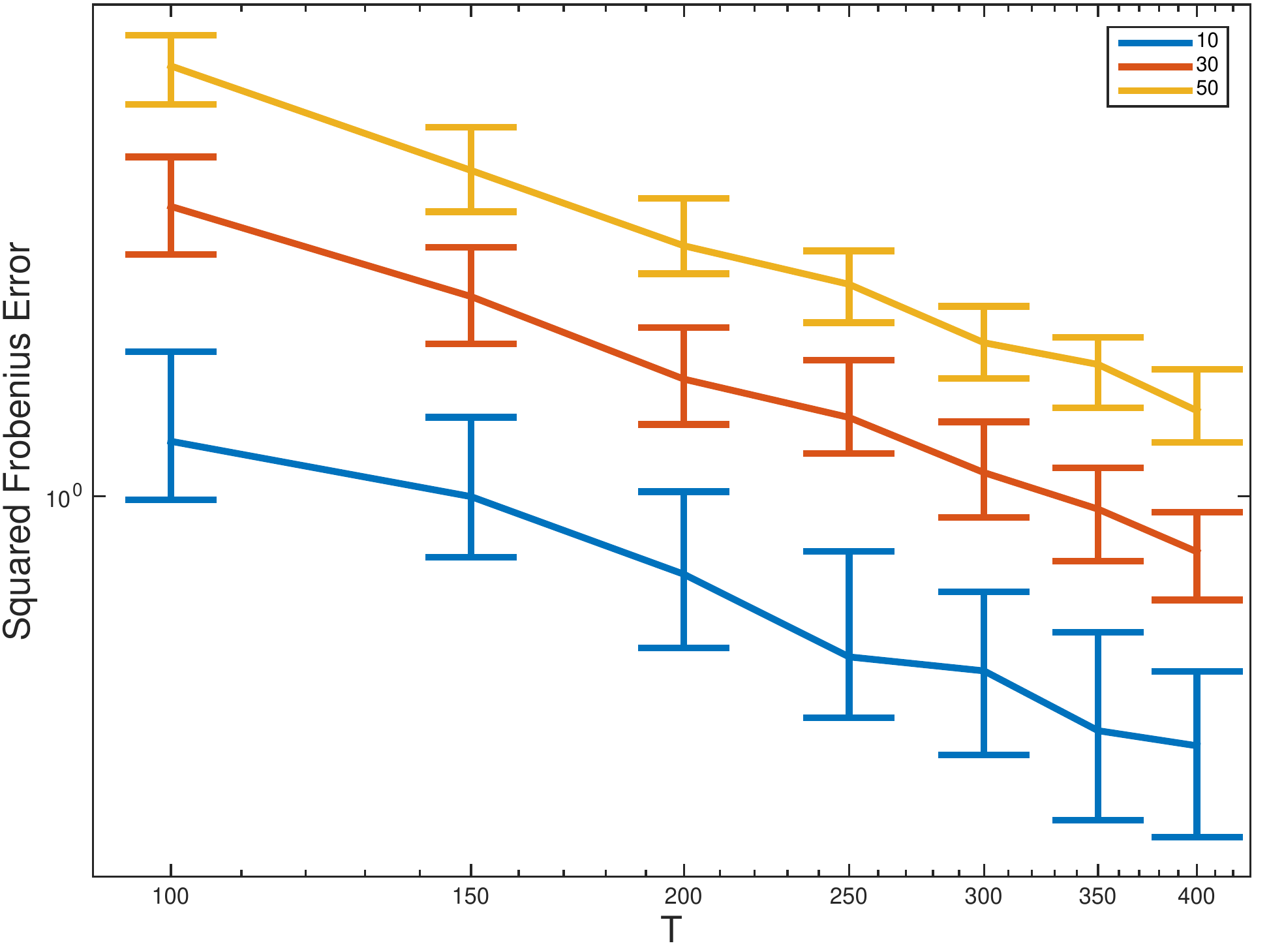}}~
    \subfloat[MSE $\cdot T$ vs $T$]{\includegraphics[height=1.5in]{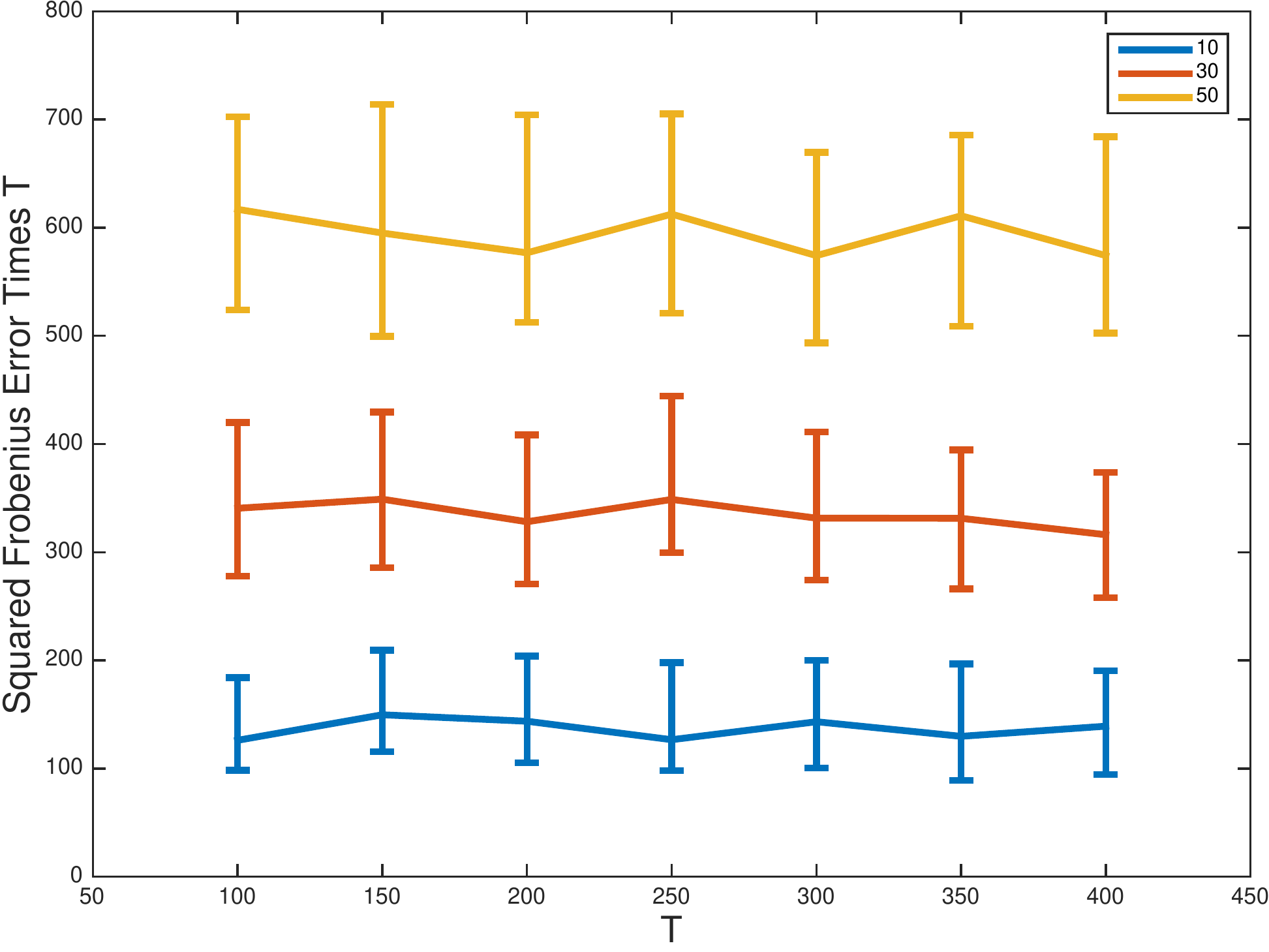}}\\
    \subfloat[MSE vs $s$]{\includegraphics[height=1.5in]{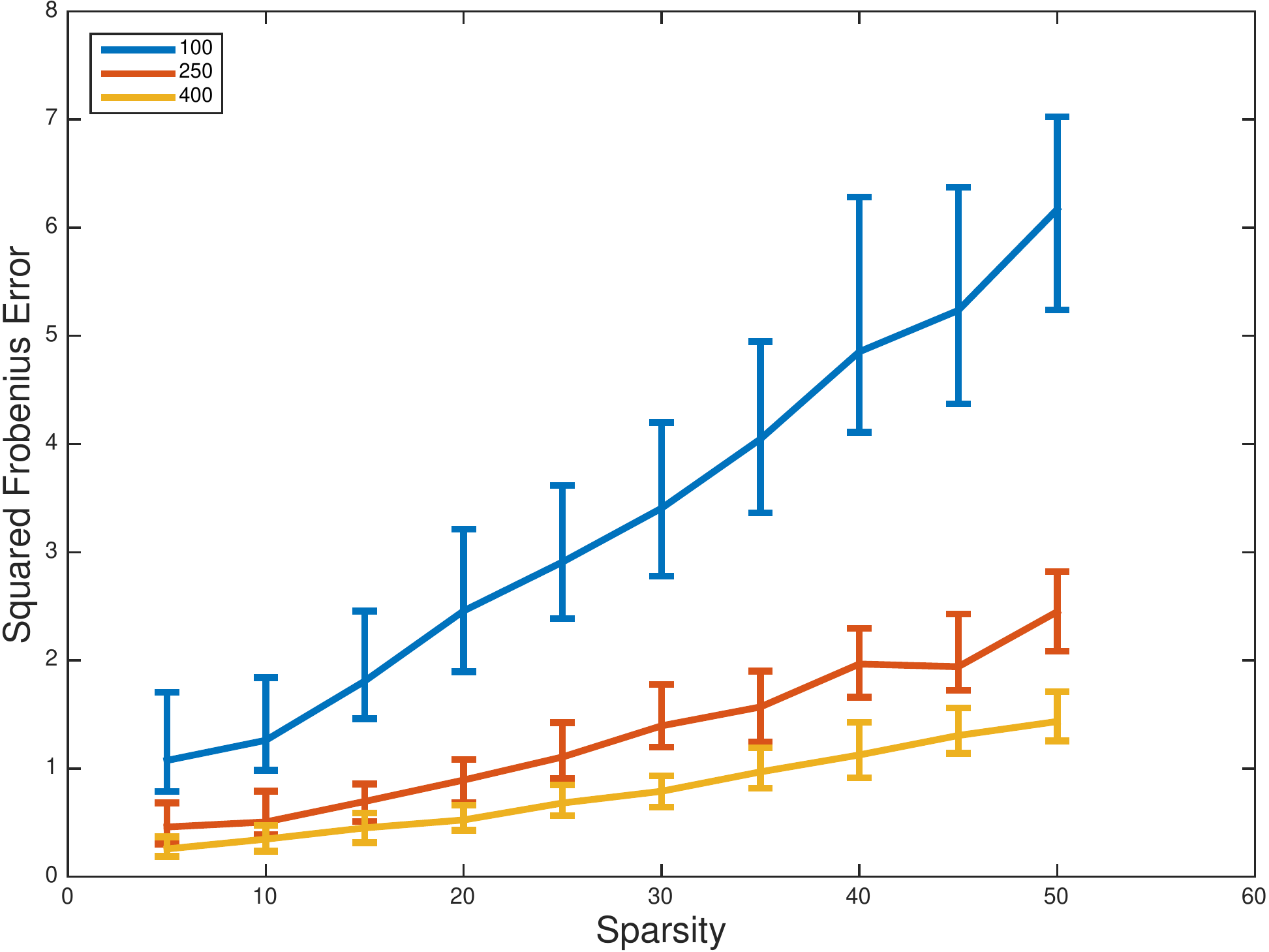}}~
    \subfloat[MSE/$s$ vs $s$]{\includegraphics[height=1.5in]{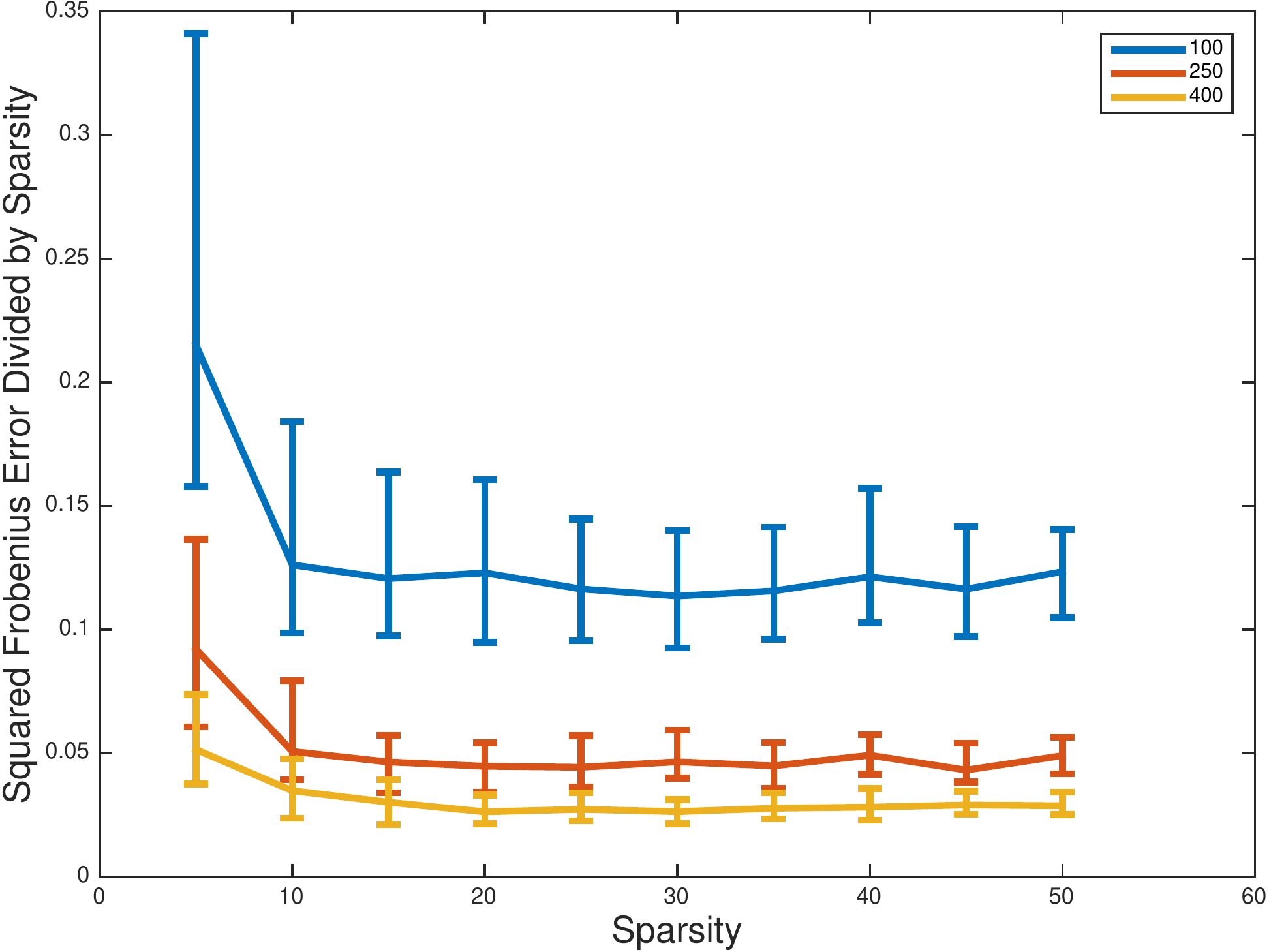}}     
     \caption{The top row of plots shows the MSE behavior over a range of $T$ values, from 100 to 400 all less than or equal to $M^2 = 400$ , where (a) is the MSE and (b) is the MSE multiplied by $T$ to show that the MSE is behaving as 1/$T$. The bottom row shows the MSE behavior over a range of $s$ values, where (c) shows MSE and (d) shows MSE divided by $s$ to show that the MSE is linear is $s$. In all plots the median value of 100 trials is shown, with error bars denoting the middle 50 percentile.}
        \label{fig:MSECurves}
\end{figure}
We validate our theoretical results with experimental results
performed on synthetically generated data using the Poisson
autoregressive process. We generate many trials of synthetic data with
known underlying parameters and then compare the estimated values. For
all trials the constant offset vector $\nu$ is set identically at 0,
and the $20\times20$ matrices $A^*$ are set such that $s$ randomly
assigned values are in the range $[-1,0]$ and with constant
$\rho = 5$. Data is then generated according the process described in
Equation~\ref{EqnModel} with the Poisson distribution. $X_0$ is chosen
as a 20 dimensional vector drawn randomly from ${\rm{Poisson}}(1)$,
then $T$ observations are used to perform the estimation. The
parameters $s$ and $T$ are then varied over a wide range of
values. For each $(s,T)$ pair 100 trials are performed, the
regularized maximum likelihood estimate $\hat{A}$ is calculated with
$\lambda = 0.1/\sqrt{T}$ and the MSE is recorded. The MSE curves are
shown in Figure~\ref{fig:MSECurves}.  Notice that the true
  values of $A^*$ are bounded by -1 and 0, but in our implementation
  we do not enforce these bounds (we set $a_{\min} = -\infty$ and
  $a_{\max} = \infty$ in Equation \ref{EqnLogLike}). While
  $a_{\min} = - \infty$ would cause the theoretical bounds to be poor,
  the theory can be applied with the smallest and largest elements of
  the matrix estimated from the unconstrained optimization. In other
  words, the theory depends on having an upper and lower bound on the
  rates, but mostly as a theoretical convenience, while the estimator
  can be computed in an unconstrained way.

\begin{figure}[t!]
    \centering
    \subfloat[Ground Truth $A^*$ Matrix]{\includegraphics[height=1.5in,width=2 in]{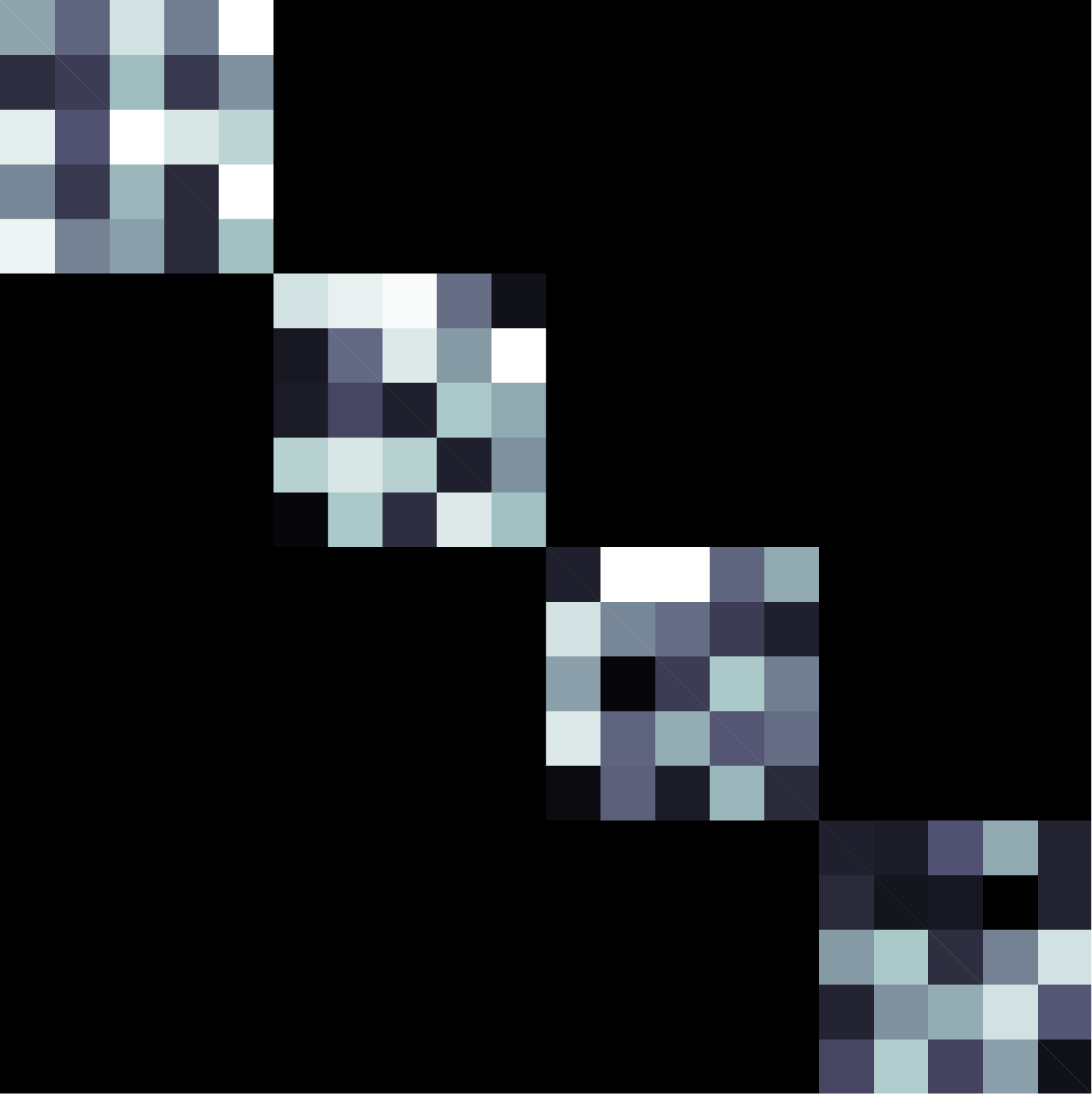}}~
    \subfloat[Estimate for $T=100$]{\includegraphics[height=1.5in,width=2 in]{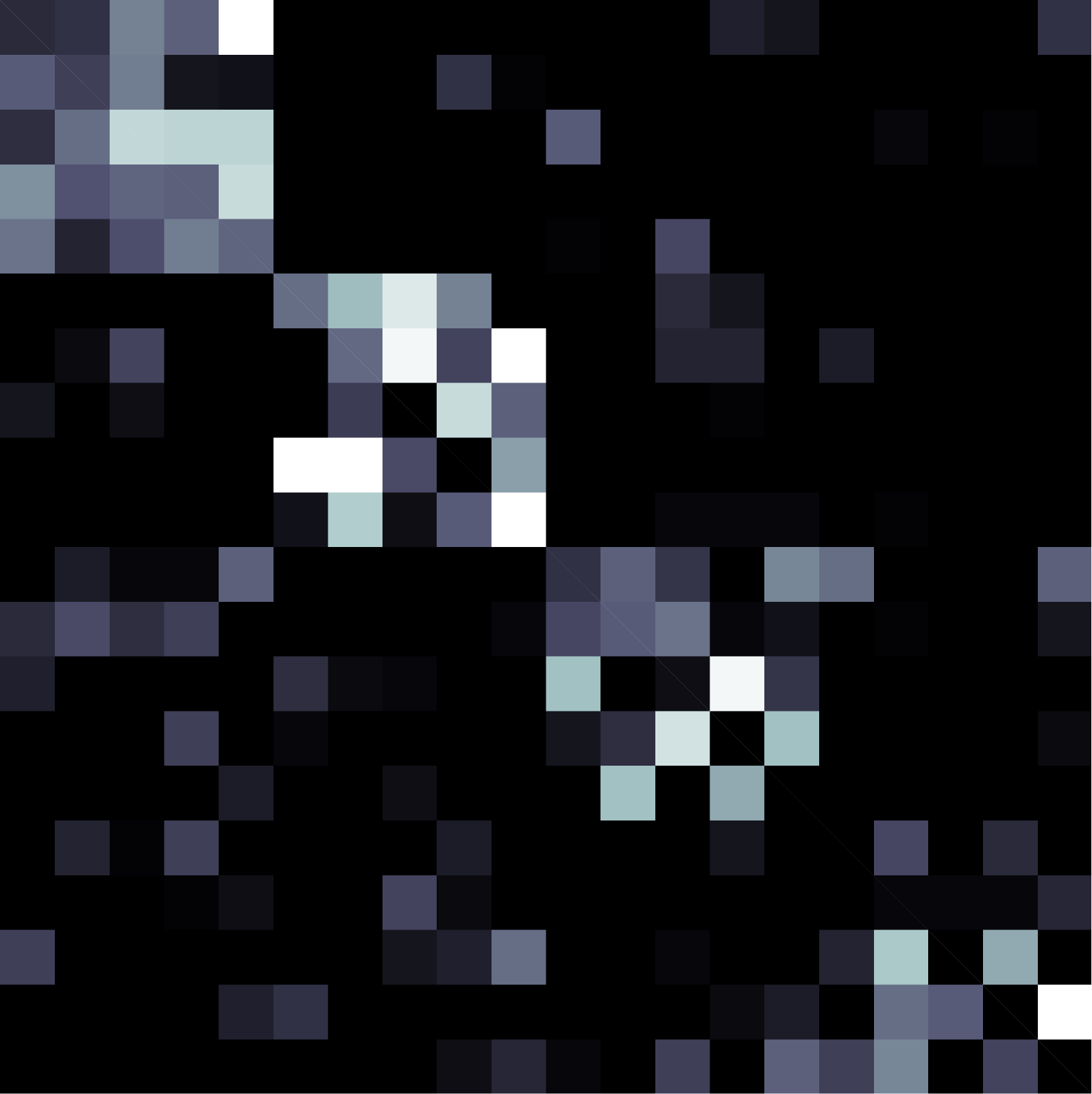}}\\
    \subfloat[Estimate for $T=316$]{\includegraphics[height=1.5in,width=2 in]{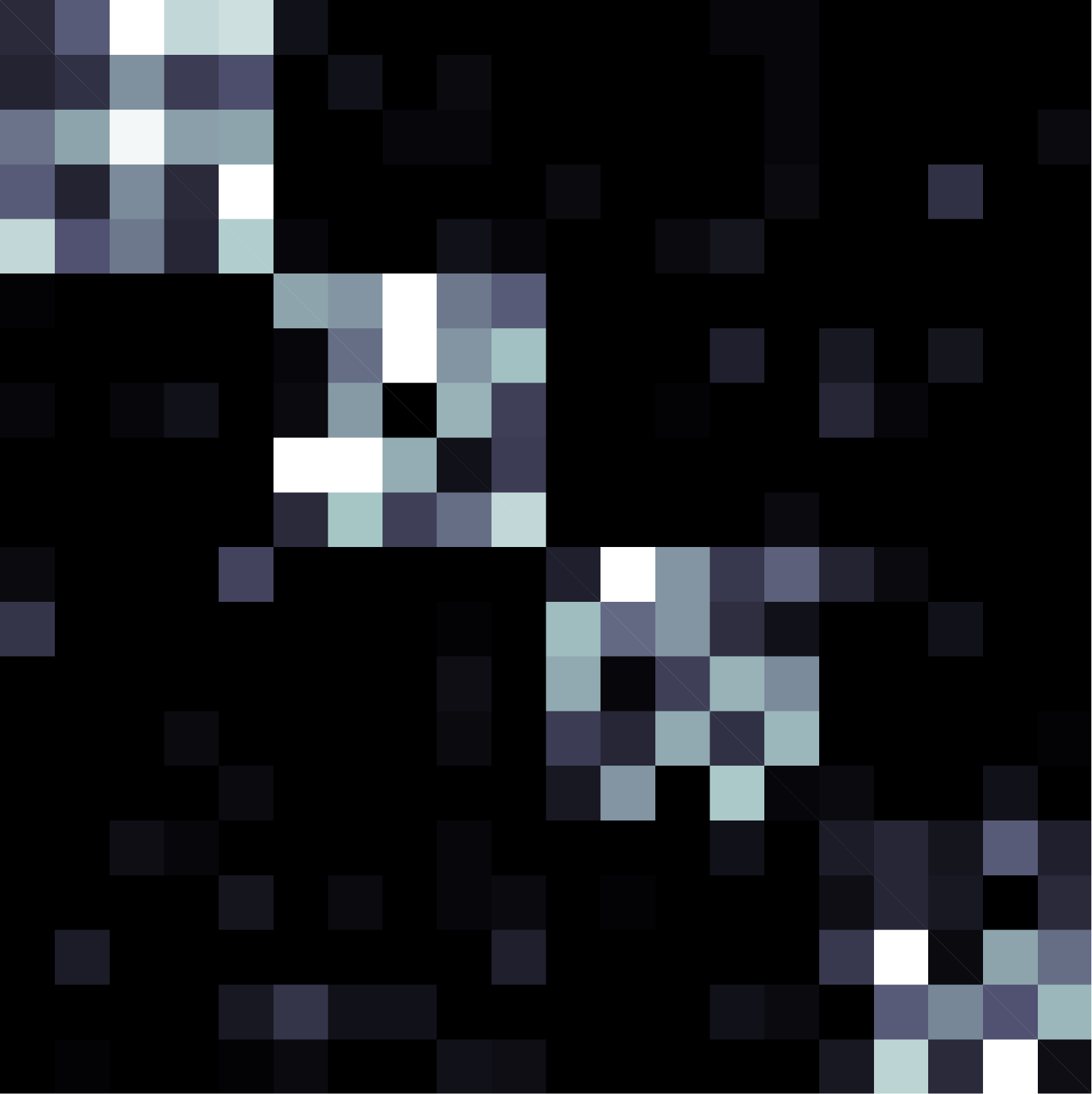}}~
    \subfloat[Estimate for $T=1000$]{\includegraphics[height=1.5in,width=2 in]{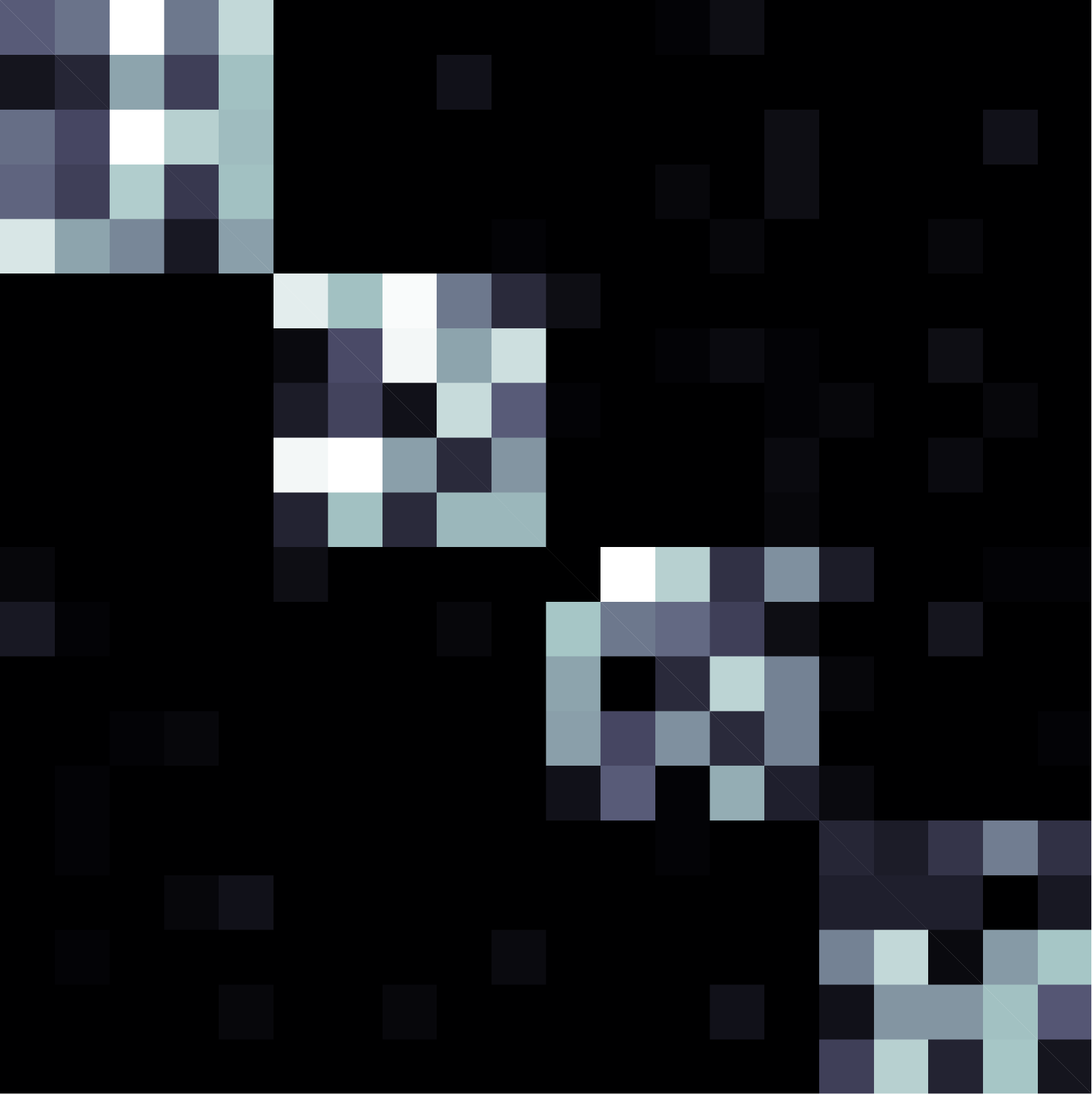}}     
     \caption{These images show the ground truth $A^*$ matrix (a) and 3 different estimates of the matrix created using increasing amounts of data. We observe that even for a relatively low amount of data we have picked out most of the support but with several spurious artifacts. As the amount of data increases, fewer of the erroneous elements are estimated. All images are scaled from 0 (dark) to -1 (bright).}
        \label{fig:RMLEstimates}
\end{figure}

We show a series of plots which compare the MSE versus increasing
behavior of $T$ and $s$, as well as comparing the behavior of
MSE$\cdot T$ and of MSE/$s$. Plotted in each figure is the
median of 100 trials for each $(s,T)$ pair, with error bars denoting
the middle 50 percentile. These plots show that setting
$\lambda$ proportional to $T^{-1}$ gives us the desired $T^{-1}$ error
decay rate. Additionally, we see that the error increases approximately
linearly in the sparsity level $s$, as predicted by the
theory. Finally, Figure~\ref{fig:RMLEstimates} shows one specific
example process and the estimates produced. The first image is the ground
truth matrix, generated to be block diagonal, in order to more easily
visualize support structure whereas in the first experiment the
support is chosen at random. One set of data is generated
using this matrix, and then estimates are constructed using the first
$T=100, 316$ and $1000$ data points. The figure shows how with more
data, the estimates become closer to the original, where much of the
error comes from including elements off the support of the true
matrix.

\begin{figure}[t!]
    \centering
    \subfloat[MSE vs. $T$]{\includegraphics[height=1.5in]{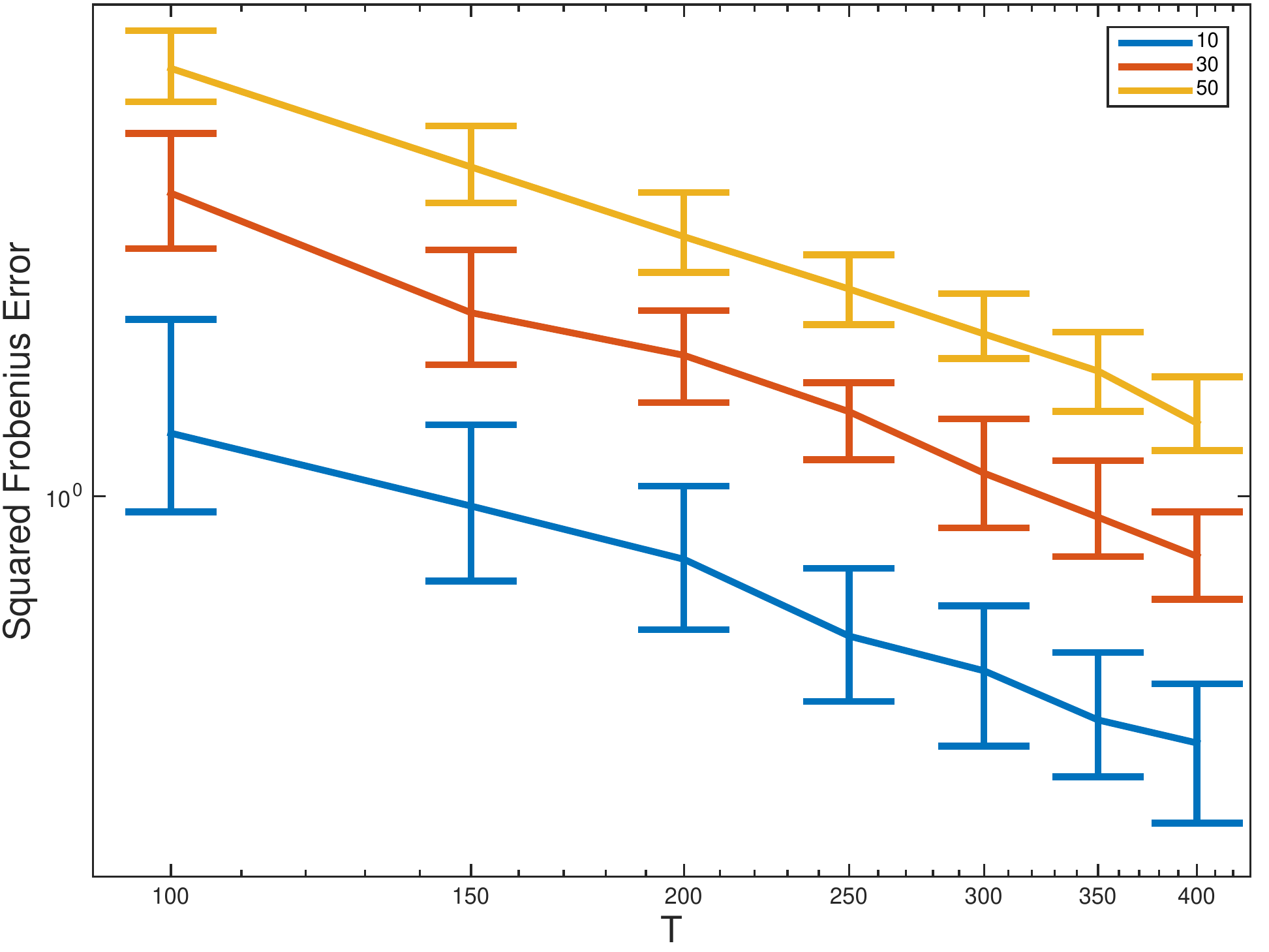}}~
    \subfloat[MSE $\cdot T$ vs $T$]{\includegraphics[height=1.5in]{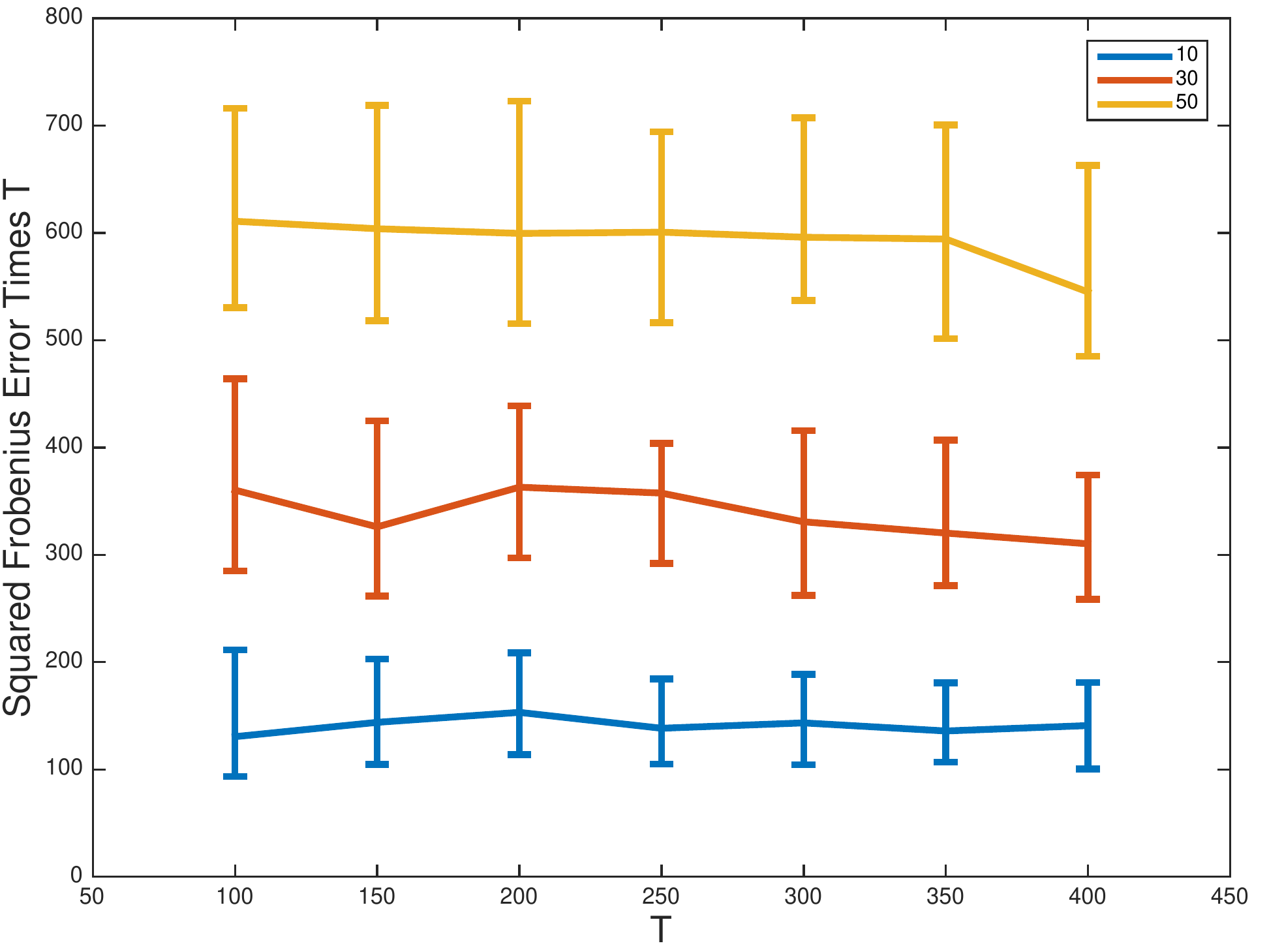}}\\
    \subfloat[MSE vs $s$]{\includegraphics[height=1.5in]{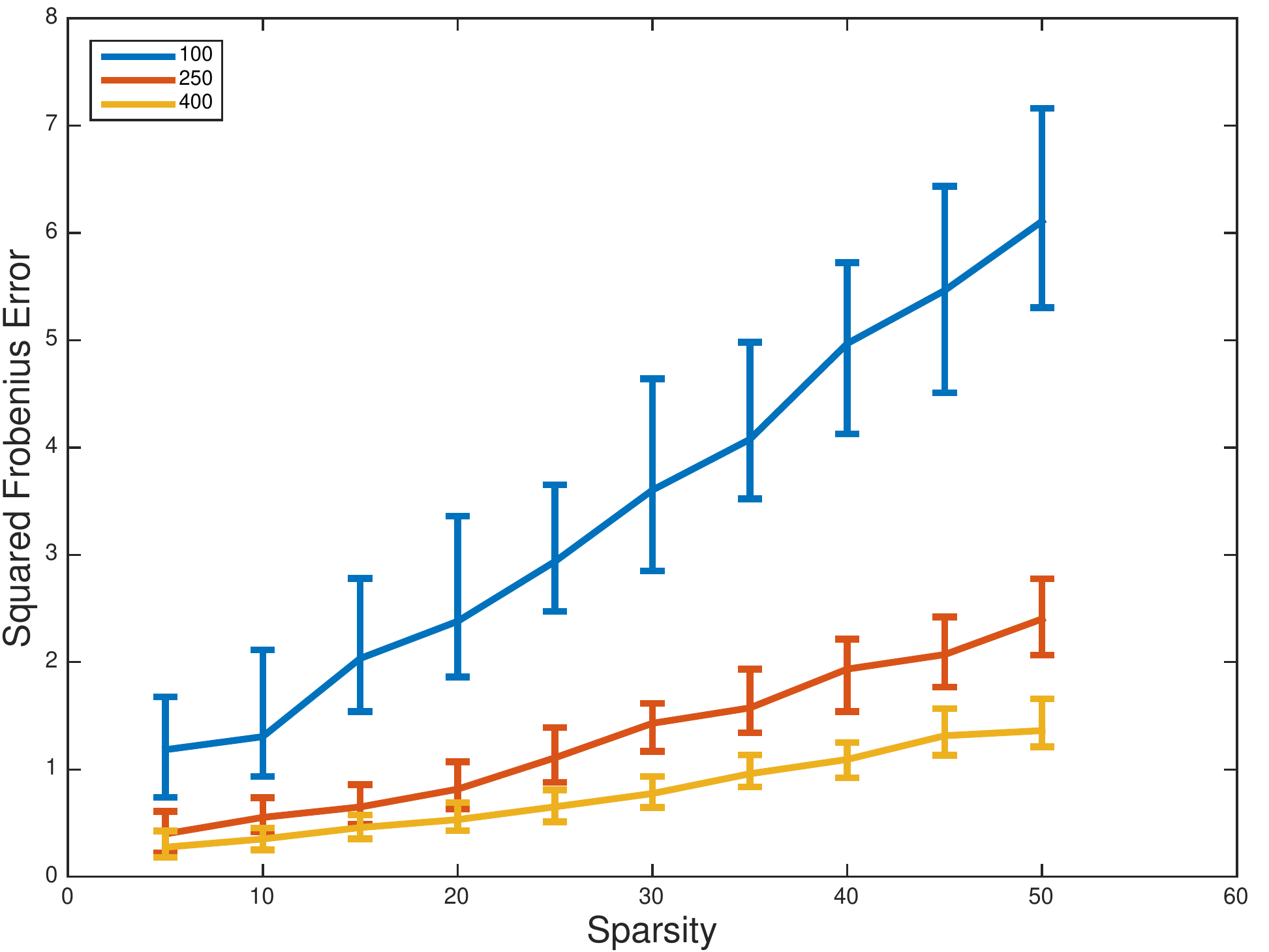}}~
    \subfloat[MSE/$s$ vs $s$]{\includegraphics[height=1.5in]{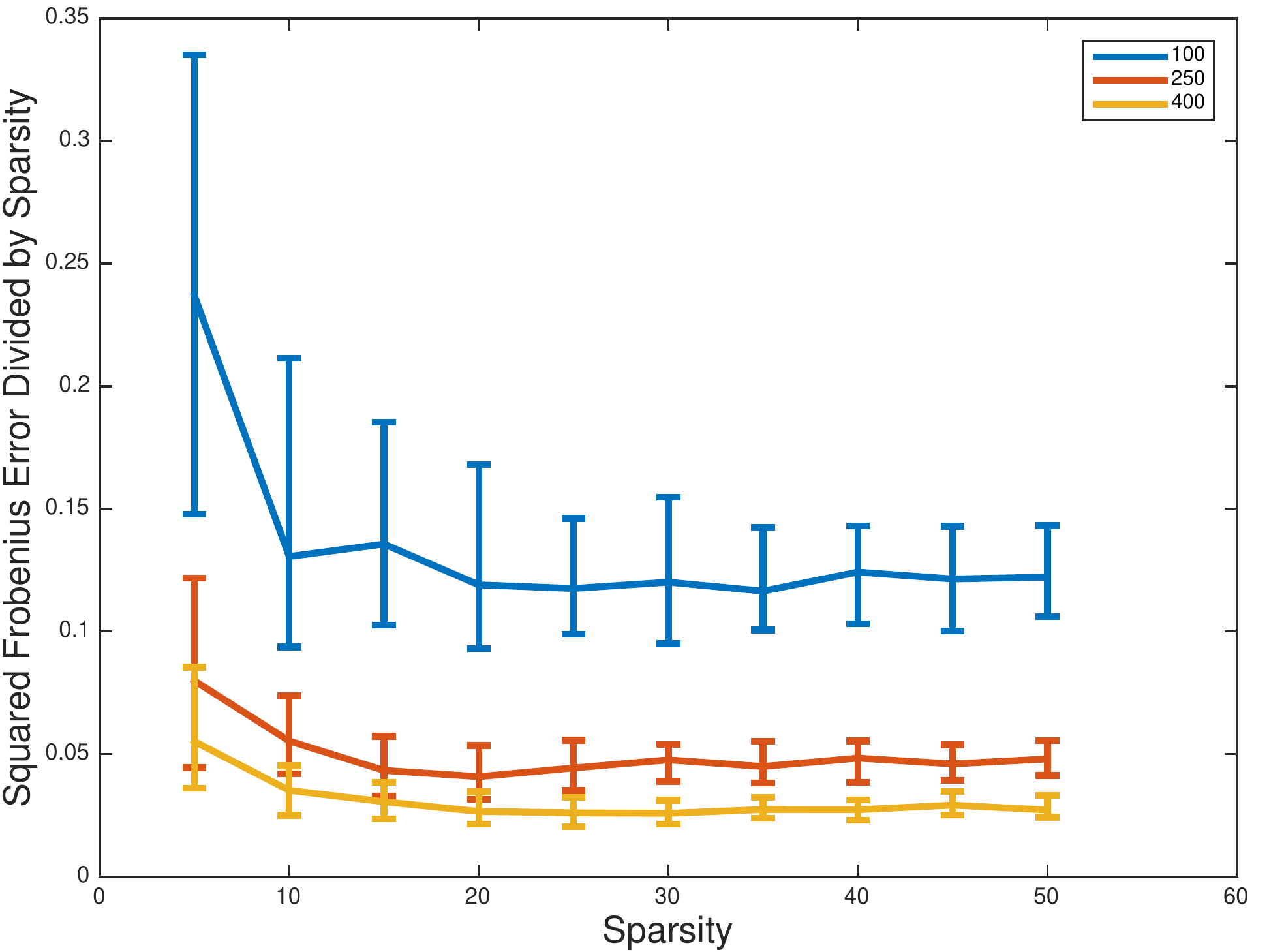}}     
     \caption{Repeat of experimental set up from Figure \ref{fig:MSECurves}, but now allowing for mixing. The top row of plots shows the MSE behavior over a widely varying range of $T$ values, from 100 to 400, where (a) is the MSE and (b) is the MSE multiplied by $T$ to show that the MSE is behaving as 1/$T$. The bottom row shows the MSE behavior over a range of $s$ values, where (c) shows MSE and (d) shows MSE divided by $s$ to show that the MSE is linear is $s$. In all plots the median value of 100 trials is shown, with error bars denoting the middle 50 percentile. Most importantly, the behavior and magnitude of errors in this plot matches the results with no mixing.}
        \label{fig:BurnIn}
\end{figure}

One important characteristic of the our results is that it does not depend on any assumptions
about the stationarity or the mixing time of the process. To show that this is truly a property of the system
and not just our proof technique, we repeat the experimental process described above, but for each set of 
observations of length $T$, we first generate 10,000 observations to allow the process to mix. In other words,
for every matrix $A$ we generate $T + 10,000$ observations, but only use the last $T$ to find the RMLE. The plots in 
figure \ref{fig:BurnIn} show the results of this experiment. The important observation is that the results both scale the same
way, and have approximately the same magnitude as the experiment when no mixing was done.

\section{Proofs}
\label{sec:proofs}
\subsection{Proof of Theorem \ref{ThmMain}}
\spro
We start the proof by making an important observation about the estimator defined in Equation \ref{EqnLogLike}: this loss function can be completely decoupled by a sum of functions on rows. Therefore we can bound the error of a single row of the RMLE and add the errors to get the final bound. For each row we use a standard method in empirical risk minimization and the definition of the minimizer of the regularized likelihood for each row:
\begin{align*}
\frac{1}{T}&\sum_{t=0}^{T-1}  Z(\nu_m + \widehat{a}_m^\T X_t ) - \widehat{a}_m^{\T}X_t \phi(X_{t+1,m}) + \lambda \|\widehat{a}_m\|_1\\
& \leq  \frac{1}{T}\sum_{t=0}^{T-1}  Z(\nu_m+a_m^{*\T}X_t) + a_m^{*\T}X_t \phi(X_{t+1,m}) + \lambda \|a^*_m\|_1.
\end{align*}
We define $\epsilon_{t,m} \triangleq \phi(X_{t+1,m}) - \mathbb{E}[\phi(X_{t+1,m})|X_t]$, which is conditionally zero mean random variable. By using a moment generating function argument, we know that $\mathbb{E}[\phi(X_{t+1,m})|X_t] = Z'(\nu_m + a_m^{*\T}X_t)$, and therefore $\phi(X_{t+1, m}) = Z'(\nu_m + a_m^{*\T}X_t) + \epsilon_{t,m}$. Hence
\begin{align*}
\frac{1}{T}&\sum_{t=0}^{T-1}Z(\nu_m+\widehat{a}_m^{\T}X_t) - \widehat{a}_m^{\T}X_t (Z'(\nu_m + a_m^{*\T}X_t) + \epsilon_{t,m}) + \lambda \|\widehat{a}_m\|_1\\
 & \leq  \frac{1}{T}\sum_{t=0}^{T-1} {Z(\nu_m+a_m^{*\T}X_t) - a_m^{*\T}X_t (Z'(\nu_m + a_m^{*\T}X_t) + \epsilon_{t,m})} + \lambda \|a^*_m\|_1.
\end{align*}
Now we use the definition of a Bregman divergence to lower bound the left hand side. An important property of Bregman divergences is that if they are induced by a strongly convex function, then the Bregman can be lower bounded by a scaled $\ell_2$ difference of its arguments. This is where our squared error term will come.
\begin{align*}
\frac{1}{T}\sum_{t=0}^{T-1} &{\left(Z(\nu_m+\widehat{a}_m^{\T}X_t)-Z(\nu_m + a_m^{*\T}X_t) - Z'(\nu_m + a_m^{*\T}X_t)(\widehat{a}_m^{\T}X_t - a_m^{*\T}X_t)\right)}\\  \leq &  \left|\frac{1}{T}\sum_{t=0}^{T-1} {\epsilon_{t,m} \Delta_m^{\T}X_t } \right| + \lambda (\|a_m^*\|_1 - \|\widehat{a}_m\|_1),
\end{align*}
where $\Delta_m = \widehat{a}_m - a_m^*$. Let $B_Z(\cdot\|\cdot)$ denote the Bregman divergence induced by $Z$. Hence
\begin{eqnarray*}
\frac{1}{T}\sum_{t=0}^{T-1} {B_Z(\nu_m+\widehat{a}_m^{\T}X_t\| \nu_m + a_m^{*\T}X_t) } & \leq & \left|\frac{1}{T}\sum_{t=0}^{T-1} {\epsilon_{t,m} \Delta_m^{\T}X_t } \right| + \lambda (\|a_m^*\|_1 -\|\widehat{a}_m\|_1).
\end{eqnarray*}
First we upper bound the right-hand side of the inequality as follows:
\begin{eqnarray*}
\frac{1}{T}\sum_{t=0}^{T-1} {B_Z(\nu_m+\widehat{a}_m^{\T}X_t\| \nu_m + a_m^{*\T}X_t) } & \leq & \left|\frac{1}{T}\sum_{t=0}^{T-1}  {\epsilon_{t,m} \Delta_m^{\T}X_t } \right| + \lambda (\|a^*_m\|_1 - \|\widehat{a}_m\|_1) \\
& = & \left|\frac{1}{T}\sum_{t=0}^{T-1} {\epsilon_{t,m} \Delta_m^{\T}X_t } \right| + \lambda (\|a^*_{m,\mathcal{S}}\|_1 - \|\widehat{a}_{m,\mathcal{S}}\|_1 - \|\widehat{a}_{m,\mathcal{S}^c}\|_1)\\
& \leq &  \left|\frac{1}{T}\sum_{t=0}^{T-1} {\epsilon_{t,m} \Delta_m^{\T}X_t } \right| + \lambda \|\Delta_{m,\mathcal{S}}\|_1 - \lambda \|{\Delta}_{m,\mathcal{S}^c}\|_1\\
& \leq & \|\Delta_m\|_1\left\|\frac{1}{T} \sum_{t=0}^{T-1} X_t \epsilon_{t,m}\right\|_\infty + \lambda \|\Delta_{m,\mathcal{S}}\|_1 - \lambda\|{\Delta}_{m,\mathcal{S}^c}\|_1.
\end{eqnarray*}
In the above, we use the defintion of $\mathcal{S}$ as the true support of $A^*$ and have used the decomposability of $\|\cdot\|_1$. The decomposability of the norm means that we have the property 
\begin{align*}
\|x\|_{1} = \|x_{\mathcal{S}}\|_{1} + \|x_{\mathcal{S}^C}\|_{1}.
\end{align*}

Note that $\left\|\frac{1}{T} \sum_{t=0}^{T-1} X_t \epsilon_{t,m}\right \|_\infty \leq \max_{1 \leq m \leq M}\left\|\frac{1}{T} \sum_{t=0}^{T-1} X_t \epsilon_{t,m}\right \|_\infty$. Under the assumption that $\max_{1 \leq m \leq M}\left\|\frac{1}{T} \sum_{t=0}^{T-1} X_t \epsilon_{t,m}\right \|_\infty \leq \lambda/2$ and by the non-negativity of the Bregman divergence on the left hand side of the inequality, we have that
\begin{align*}
0 \leq \frac{\lambda}{2} \|\Delta_m\|_1 + \lambda \|\Delta_{m,\mathcal{S}}\|_1 - \lambda \|\Delta_{m,\mathcal{S}^c}\|_1.
\end{align*}
Using the decomposability of the $\ell_1$ norm, this inequality implies that for all rows $1 \leq m \leq M$, we have that $\|\Delta_{m,\mathcal{S}^c}\|_1 \leq 3 \|\Delta_{m,\mathcal{S}}\|_1$. Since $\|\Delta_{m,\mathcal{S}^c}\|_1 \leq 3 \|\Delta_{m,\mathcal{S}}\|_1$, $ \|\Delta_{m}\|_1 \leq 4 \|\Delta_{m,\mathcal{S}}\|_1$ and consequently
$$
\|\Delta_{m}\|_1 \leq 4\sum_{j \in \mathcal{S}} |\Delta_{m,j}| \leq 8 \rho_m \tilde{a}
$$
where the final inequality follows since $| \Delta_{m,j} | \leq 2 \tilde{a}$ for all $j$. Using this inequality and the fact that $\|a^*_m\|_1 \leq \rho_m \tilde{a}$ implies that $\|\hat{a}_m\|_1 \leq 9 \rho_m \tilde{a}$, and therefore for all $t \in \mathcal{T}$ the range of both $\nu_m + a_m^{*\top}X_t$ and $\nu_m + \hat{a}_m^\top X_t$ are in $[-\tilde{\nu} - 9 \rho \tilde{a}, \tilde{\nu} + 9 \rho \tilde{a}].$ 

Now to lower bound the Bregman divergence in terms of the Frobenius norm, we use the first condition of Assumption \ref{AsLearnable}. Inherently, the RMLE will admit estimates which should converge to the true matrix $A^*$ under a Bregman divergence induced by the log-partition function, but we are interested in convergence of the Frobenius norm. Therefore, to convert from one to the other, we require the log-partition function to be strongly convex. This issue is side-stepped in the Gaussian noise case, due to the fact that the Bregman in question would identically be the Frobenius norm. By Assumption \ref{AsLearnable}, $Z$ is $\sigma$-strongly convex, and therefore on $\mathcal{T}$ it is true that $B_Z(\nu + \widehat{a}_m^{\T}X_t\| \nu_m + a_m^{*\T}X_t) \geq \frac{\sigma}{2} (\Delta_m^{\T}X_t)^2$ and $B_Z(\nu_m+\widehat{a}_m^{\T}X_t\| \nu_m + a_m^{*\T}X_t) \geq 0$ on the rest of the time indices. 

Therefore
\begin{eqnarray*}
\frac{1}{T}\sum_{t=0}^{T-1} {B_Z(\nu_m+\widehat{a}_m^{\T}X_t\| \nu_m + a_m^{*\T}X_t) }
& \leq & \frac{\lambda}{2}\|\Delta_m\|_1 + \lambda \|\Delta_{m,\mathcal{S}}\|_1 - \lambda\|{\Delta}_{m,\mathcal{S}^c}\|_1,
\end{eqnarray*}
implies
\begin{eqnarray*}
\frac{\sigma}{2T}\sum_{t \in \mathcal{T}} {(\Delta_m^{\T}X_t)^2} 
& \leq & \frac{\lambda}{2}\|\Delta_m\|_1 + \lambda \|\Delta_{m,\mathcal{S}}\|_1 - \lambda\|{\Delta}_{m,\mathcal{S}^c}\|_1.
\end{eqnarray*}
Define $\|\Delta_m\|_T^2 = \frac{1}{T} \sum_{t \in \mathcal{T}}{(\Delta_m^\top X_t)^2}$ for any $\Delta \in \mathbb{R}^{M \times M}$, then we have the bound:

 \begin{eqnarray*}
\frac{\sigma}{2} \| \Delta_m\|_T^2 & \leq & \frac{\lambda}{2} \|\Delta_m\|_{1}+ \lambda \|\Delta_{m,\mathcal{S}}\|_{1} - \lambda \|\Delta_{m,\mathcal{S}^c}\|_{1} \leq \frac{3\lambda}{2} \|\Delta_{m,\mathcal{S}}\|_{1}.
\end{eqnarray*}

Therefore we can define the cone on which the vector $\Delta_m$ must be defined:
\begin{equation*}
\mathcal{B}_{m,\mathcal{S}} := \{\Delta \in [a_{\min} - a_{\max},a_{\max}-a_{\min}]^{M}\;|\;  \|\Delta_{m,\mathcal{S}^c}\|_1 \leq 3 \|\Delta_{m,\mathcal{S}}\|_1\},
\end{equation*} 
and restrict ourselves to studying properties of vectors in that set. Since $\|\Delta_{m,\mathcal{S}}\|_{1} \leq \sqrt{\rho_m} \| \Delta_m\|_2$ where $\rho_m$ is the number of non-zeros of $a^*_m$, we have that
\begin{eqnarray}
\| \Delta_m\|_T^2 & \leq & \frac{3}{\sigma} \lambda \sqrt{\rho_m} \| \Delta_m\|_2 = \delta_m \| \Delta_m\|_2, \label{eq:genBound}
\end{eqnarray}
where $\delta_m \triangleq \frac{3}{\sigma} \lambda \sqrt{\rho_m}$. 
Now we consider three cases: if $\|\Delta_m\|_T \geq \| \Delta_m\|_2$, then $\max(\| \Delta_m\|_T, \| \Delta_m\|_2) \leq \delta_m$. On the other hand if $\|\Delta_m\|_T \leq \| \Delta_m \|_2$ and $\|\Delta_m\|_2 \leq \delta_m$, then $\max(\| \Delta_m\|_T, \| \Delta_m\|_2) \leq \delta_m $. 

Hence the final case we need to consider is $\|\Delta_m\|_T \leq \| \Delta_m \|_2$ and $\|\Delta_m\|_2 \geq \delta_m$. Now we follow a similar proof technique to that used in Raskutti et al.~\cite{RasWaiYu12} adapted to dependent sequences, to understand this final scenario. Let us define the following set:
\begin{eqnarray}
\mathcal{B}_m(\delta_m) := \{ \Delta_m \in \mathcal{B}_{m,{\mathcal{S}}}\; |\; \|\Delta_m\|_2 \geq \delta_m \}. \label{EqHypSet}
\end{eqnarray}
Further, let us define the alternative set:
\begin{equation}
\mathcal{B}'_m(\delta_m) := \{ \Delta_m \in \mathcal{B}_{m,\mathcal{S}}\; |\;\|\Delta_m\|_2 = \delta_m \}. \label{EqAltSet}
\end{equation}
We wish to show that for $\Delta_m \in \mathcal{B}_m(\delta_m),$ we have
$\|\Delta_m\|_T^2 \geq \kappa \|\Delta_m\|_2^2$ for some
$\kappa \in (0,1)$ with high probability, and therefore Equation
\ref{eq:genBound} would imply that
$\max(\|\Delta_m\|_T,\|\Delta_m\|_2) \leq \delta_m/\kappa$. We claim
that it suffices to show that
$\|\Delta_m\|_T^2 \geq \kappa\|\Delta_m\|_2^2$ is true on
$\mathcal{B}_m'(\delta_m)$ with high probability. In particular, given
an arbitrary non-zero $\Delta_m \in \mathcal{B}_m(\delta_m)$, consider
the re-scaled vector
$\tilde{\Delta}_m = \frac{\delta_m}{\|\Delta_m\|_2} \Delta_m$. Since
$\Delta_m \in \mathcal{B}_m(\delta_m)$, we have
$\tilde{\Delta}_m \in \mathcal{B}_m(\delta_m)$ and
$\|\tilde{\Delta}_m\|_2 = \delta_m$ by construction. Together, these
facts imply $\tilde \Delta_m \in \mathcal{B}'_m(\delta_m)$. Furthermore, if
$\|\tilde\Delta_m\|_T^2 \geq \kappa \|\tilde\Delta_m\|_2^2$ is true,
then $\|\Delta_m\|_T^2 \geq \kappa \|\Delta_m\|_2^2$ is also
true. Alternatively if we define the random variable
$\mathcal{Z}_T(\mathcal{B}'_m) = \sup_{\Delta_m \in
  \mathcal{B}'_m(\delta_m)} \{\delta_m^2 - \| \Delta_m\|_T^2 \}$,
then it suffices to show that
$\mathcal{Z}_T(\mathcal{B}'_m) \leq (1-\kappa)\delta_m^2$.

For this step we use some recent concentration bounds~\cite{VDGMartingale} and empirical process techniques~\cite{RakhlinTewari} for martingale random variables. Recall that the empirical norm is $\|\Delta_m\|_T^2 = \frac{1}{T}\sum_{t \in \mathcal{T}}{(\Delta_m^T X_t)^2}$. Further let $(t_i)_{i=1}^{|\mathcal{T}|}$ denote the indices in $\mathcal{T}$. Next we define the conditional expectation
$$
Y_T := \frac{1}{T}\sum_{i = 1}^{|\cal T|}\mathbb{E}\big[(\Delta_m^TX_{t_i} )^2|X_{t_1}, X_{t_2},\ldots,X_{t_i-1} \big]. 
$$
Then we have
$$
\mathcal{Z}_T(\mathcal{B}'_m) = \sup_{\Delta_m \in \mathcal{B}'_m(\delta_m)} \{\delta_m^2 - \| \Delta_m\|_T^2\} \leq \sup_{\Delta_m \in \mathcal{B}'_m(\delta_m)} \{\delta_m^2 -  Y_T\} + \sup_{\Delta_m \in \mathcal{B}'_m(\delta_m)} \{Y_T -  \|\Delta_m\|_T^2\}.
$$
To bound the first quantity, $\sup_{\Delta_m \in \mathcal{B}'_m(\delta_m)} \{\delta_m^2 -  Y_T\}$, we first note that
$$
\sup_{\Delta_m \in \mathcal{B}'_m(\delta_m)} \{\delta_m^2 -  Y_T\} \leq \delta_m^2 - \delta_m^2\omega = (1 -\omega)\delta_m^2
$$
by Assumption~\ref{AsLearnable} and the fact that $\|\Delta_m\|_2^2 =
\delta_m^2$ since $\Delta_m \in \mathcal{B}'_m(\delta_m)$. Thus
$$
\mathcal{Z}_T(\mathcal{B}'_m) \leq (1 - \omega)\delta_m^2  + \sup_{\Delta_m \in \mathcal{B}'_m(\delta_m)} \{Y_T -  \|\Delta_m\|_T^2\}.
$$
Now we focus on bounding $\sup_{\Delta_m \in \mathcal{B}'_m(\delta_m)} \{Y_T -  \|\Delta_m\|_T^2\}$. First, we use a martingale version of the bounded difference inequality using Theorem 2.6 in ~\cite{VDGMartingale} (see Appendix~\ref{App:Emp}):
$$
\sup_{\Delta_m \in \mathcal{B}'_m(\delta_m)} \{Y_T -  \|\Delta_m\|_T^2\} \leq \mathbb{E}[\sup_{\Delta_m \in \mathcal{B}'_m(\delta_m)} \{Y_T -  \|\Delta_m\|_T^2\}] + \frac{\omega \delta_m^2}{4},
$$
with high probability. Recall that on $\mathcal{T}$, we have $0\leq (\Delta_m^\top X_t)^2 \leq \|\Delta_m\|_1^2 \|X_t\|_\infty^2 \leq U^2 \|\Delta_m\|_1^2$. Because $\Delta_m \in \mathcal{B}'_m(\delta_T)$, it is true that $\|\Delta_m\|_1 \leq 4 \|\Delta_{m,\mathcal{S}}\|_1$. We then use the the relationship between the $\ell_1$ and $\ell_2$ norms to say $\|\Delta_{m,\mathcal{S}}\|_1 \leq \sqrt{\rho_m} \|\Delta_{m,\mathcal{S}}\|_2 \leq \sqrt{\rho_m} \|\Delta_m\|_2$ where $\rho_m$ is the number of non-zeros in the $m^{th}$ row of the true matrix $A^*$. Putting these together means $(\Delta_m^\top X_t)^2 \leq 16 U^2 \rho_m \delta_m^2$. 
In particular, we apply Theorem~\ref{ThmMcDiarmid} in Appendix~\ref{App:Emp} with $Z_T = \sup_{\Delta_m \in \mathcal{B}'_m(\delta_m)} \{Y_T -  \|\Delta_m\|_T^2\}$, $a = \frac{\omega \delta_m^2}{4}$, $L_t = - \frac{16 U^2 \rho_m \delta_m^2}{T}$ and $U_t = \frac{16 U^2 \rho_m \delta_m^2}{T}$, and therefore $C_T^2 = \frac{32^4 U^4 \rho_m^2 \delta_m^4}{T}$. Therefore, applying Theorem~\ref{ThmMcDiarmid}
$$
\sup_{\Delta_m \in \mathcal{B}'_m(\delta_m)} \{Y_T -  \|\Delta_m\|_T^2\} \leq \mathbb{E}[\sup_{\Delta_m \in \mathcal{B}'_m(\delta_m)} \{Y_T -  \|\Delta_m\|_T^2\}] + \frac{\omega \delta_m^2}{4},
$$
with probability at least $1 - \exp(-\frac{2T}{32^4 U^4 \rho_m^2})$. Since $T \geq 32^4 U^4 \rho_m^2 \log(M)$, the above statement holds with probability at least $1 - \frac{1}{M^2}$. Hence
$$
\mathcal{Z}_T(\mathcal{B}'_m) \leq (1 - \omega)\delta_m^2  + \frac{\omega \delta_m^2}{4} + \mathbb{E}[\sup_{\Delta_m \in \mathcal{B}'_m(\delta_m)} \{Y_T -  \|\Delta_m\|_T^2\}].
$$
Now we bound $\mathbb{E}[\sup_{\Delta_m \in \mathcal{B}'_m(\delta_m)} \{Y_T -  \|\Delta_m\|_T^2\}]$. Here  we use a recent symmetrization technique adapted for martingales in~\cite{RakhlinTewari}. To do this, we introduce the so-called \emph{sequential Rademacher complexity} defined in ~\cite{RakhlinTewari}. Let $(\epsilon_t)_{t=1}^T$ be independent Rademacher random variables, that is $\mathbb{P}(\epsilon_t = +1 ) = \mathbb{P}(\epsilon_t = -1 ) = \frac{1}{2}$. For a function class $\mathcal{F}$, the sequential Rademacher complexity $\mathcal{R}_T(\mathcal{F})$ is:
$$
\mathcal{R}_T(\mathcal{F}) := \sup_{X_1, X_2,\ldots,X_T} \mathbb{E}\biggr[\sup_{f \in \mathcal{F}}\frac{1}{T}\sum_{t=1}^{T}{\epsilon_t f(X_t(\epsilon_1, \epsilon_2,\ldots,\epsilon_{t-1}) )} \biggr].
$$  
Note here that $X_t$ is a function of the previous independent random variables $(\epsilon_1, \epsilon_2,\ldots,\epsilon_{t-1})$. Using Theorem 2 in ~\cite{RakhlinTewari} (also stated Appendix~\ref{App:Emp}) with $f(X_t) = (\Delta_m^TX_t)^2$ and noting that even though we use the index set $\mathcal{T}$, $(X_t)_{t \in \mathcal{T}}$ is still a martingale, it follows that:
$$
\mathbb{E}\left[\sup_{\Delta_m \in \mathcal{B}'_m(\delta_m)} \{Y_T -
\|\Delta_m\|_T^2\}\right] \leq 2 \sup_{X_{t_1},
  X_{t_2},\ldots,X_{|\mathcal{T}|}} \mathbb{E}\biggr[\sup_{\Delta_m
  \in \mathcal{B}'_m(\delta_m)}\frac{1}{T}\sum_{i=1}^{|\cal T|}{\epsilon_{t_i} (\Delta_m^T X_{t_i})^2} \biggr].
$$
Additionally since  $|\Delta_m^\top X_t| \leq 4 U \sqrt{\rho_m} \delta_m$ by the argument above and using the symmetry of Rademacher random variables
\begin{eqnarray*}
\mathbb{E}\left[\sup_{\Delta_m \in \mathcal{B}'_m(\delta_m)} \{Y_T -
  \|\Delta_m\|_T^2\}\right] & \leq & 2 \sup_{X_1, X_2,\ldots,X_{|\cal
                                     T|}}
                                     \mathbb{E}\biggr[\sup_{\Delta_m
                                     \in
                                     \mathcal{B}'_m(\delta_m)}\frac{1}{T}\sum_{i=1}^{|\cal
                                     T|}{\epsilon_{t_i} \Delta_m^T X_{t_i} |\Delta_m^T X_{t_i}| } \biggr] \\
& \leq & 8 U \sqrt{\rho_m} \delta_m \sup_{X_1, X_2,\ldots,X_{|\cal
         T|}} \mathbb{E}\biggr[\sup_{\Delta_m \in
         \mathcal{B}'_m(\delta_m)}\frac{1}{T}\sum_{i=1}^{|\cal T|}{\epsilon_{t_i} \Delta_m^T X_{t_i} } \biggr]
\end{eqnarray*}

The final step is to upper bound the sequential Rademacher
  complexity ${\cal R}_T = \mathbb{E} [\sup_{\Delta \in \mathcal{B}'_m(\delta_m)} \frac{1}{T} \sum_{i=1}^{|\cal T|} {\epsilon_{t} \Delta_m^{\T}X_{t_i}}]$ where $X_{t_i}$ is a function of $(\epsilon_1, \epsilon_2,\ldots,\epsilon_{t_i-1})$. Clearly:
\begin{eqnarray*}
\frac{1}{T} \sum_{i=1}^{|\cal T|} {\epsilon_{t} \Delta_m^{\T}X_{t_i} }
& \leq & \left \|\frac{1}{T} \sum_{i=1}^{|\cal T|}{\epsilon_t X_{t_i} }\right\|_\infty \|\Delta_m\|_{1}.
\end{eqnarray*}
Because $\Delta_m \in \mathcal{B}'_m(\delta_m)$ we have $\|\Delta_m\|_{1} = \|\Delta_{m,\mathcal{S}}\|_{1} + \|\Delta_{m,\mathcal{S}^c}\|_{1} \leq 4 \|\Delta_{m,\mathcal{S}}\|_{1}$ and $\|\Delta_{m,\mathcal{S}}\|_{1} \leq \sqrt{\rho_m} \|\Delta_{m,\mathcal{S}}\|_2 \leq \sqrt{\rho_m} \|\Delta_m\|_2 = \sqrt{\rho_m}\delta_m$. 
\begin{eqnarray*}
\mathbb{E}\left[\sup_{\Delta_m \in \mathcal{B}'_m(\delta_m)} \{Y_T -
  \|\Delta_m\|_T^2\}\right] & \leq & 8 U \sqrt{\rho_m} \delta_m \sup_{X_1,
                              X_2,\ldots,X_{|\cal T|}} \mathbb{E}\biggr[\sup_{\Delta_m \in \mathcal{B}'_m(\delta_m)}\frac{1}{T}\sum_{i=1}^{|\cal T|}{\epsilon_{t_i} \Delta_m^T X_{t_i} } \biggr]\\
& \leq & 8 U \sqrt{\rho_m} \delta_m \sup_{X_1,
         X_2,\ldots,X_{|\cal T|}}\left\|\frac{1}{T} \sum_{i=1}^{|\cal T|}\epsilon_{t_i} X_{t_i}(\epsilon_1,\ldots,\epsilon_{t_i-1})
         \right\|_{\infty} \sup_{\Delta_m \in
         \mathcal{B}'_m(\delta_m)} \|\Delta_m\|_1\\
& \leq & 32 U^2 \rho_m \delta_m^2 \sup_{X_1, X_2,\ldots,X_{|\cal T|}}\left\|\frac{1}{T} \sum_{i=1}^{|\cal T|}\epsilon_{t_i} X_{t_i}(\epsilon_1,\ldots,\epsilon_{t_i-1}) \right\|_{\infty}.
\end{eqnarray*}
Finally, we use Lemma~\ref{LemRadComp} applied to the index set $\cal T$:
\begin{eqnarray*}
\mathbb{E}\left[\sup_{\Delta_m \in \mathcal{B}'_m(\delta_m)} \{Y_T -
  \|\Delta_m\|_T^2\}\right] & \leq & 32 U^2 \rho_m \delta_m^2
                                     \sup_{X_1, X_2,\ldots,X_{|\cal
                                     T|}}\left\|\frac{1}{T}
                                     \sum_{i=1}^{|\cal
                                     T|}\epsilon_{t_i}
                                     X_{t_i}(\epsilon_1,\ldots,\epsilon_{t_i-1})
                                     \right \|_{\infty}\\
& \leq & 128 U^4 \rho_m \delta_m^2 \frac{\log(MT)}{\sqrt{T}},
\end{eqnarray*}
with probability at least $1 - \frac{1}{(MT)^2}$. Now if we set $T \geq \frac{256^2 U^8 \rho_m^2 \log^2(MT)}{\omega^2}$,
\begin{eqnarray*}
\mathbb{E}\left[\sup_{\Delta_m \in \mathcal{B}'_m(\delta_m)} \{Y_T -  \|\Delta_m\|_T^2\}\right] & \leq & \frac{\omega \delta_m^2}{4}
\end{eqnarray*}
with probability $1 - (MT)^{-2}$.

Overall this tells us that on the set $\mathcal{B}'_m(\delta_m)$ we have that $\|\Delta_m\|_T^2 \geq \frac{3\omega}{4} \|\Delta_m\|_2^2$ with high probability. Now we return to the main proof. After considering all three cases that can follow from \ref{eq:genBound}, we have
\begin{equation*}
\max(\|\Delta_m\|_2^2, \|\Delta_m\|_T^2) \leq \frac{144}{\sigma^2\omega^2 \xi^2} \rho_m \lambda^2 
\end{equation*}
with probability at least $1-\exp(\frac{c'\rho_m}{\omega^2} \log(2M)-\frac{c\omega^2T}{\rho_m^2}  )$, which bounds the error accrued on any single row, as a function of the sparsity of the true row. Combining, to get an overall error yields,
\begin{equation*}
\|\widehat{A} - A^* \|_F^2 \leq \frac{144}{\sigma^2 \omega^2 \xi^2} \lambda^2 \sum_{m=1}^M \rho_m = \frac{144}{\sigma^2 \omega^2 \xi^2} \lambda^2 s
\end{equation*}
with probability at least 
\begin{equation}
\label{eq:ProbBound}
1 - \exp\left(\log(M) + \frac{c'\rho}{\omega^2} \log(2M) - \frac{c\omega^2 T}{\rho^2}\right)
\end{equation}.
\fpro

\subsection{Proof of Theorem \ref{ThmBern}}
\subsubsection{Part 1}
\spro
The matrix $\Gamma_t$ can be expanded as
\begin{align*}
\mathbb{E}[X_{t}X_{t}^\top|X_{{t-1}}] =  \mathbb{E}[X_{t}|X_{t-1}]\mathbb{E}[X_{t}|X_{t-1}]^\top + {\rm{Diag}}({\rm{Var}}(X_{t}|X_{t-1}))
\end{align*}
Thus $\Gamma_t$ has two parts, one is the outer product of a vector with itself, and the second is a diagonal matrix. Therefore, the smallest eigenvalue will be lower bounded by the smallest element of the diagonal matrix, because the outer product matrix will always be positive semi-definite with smallest eigenvalue equal to 0. Using properties of the Bernoulli distribution, the conditional variance is explicitly given as $(2 + \exp(\nu + A^* X_{t-1}) + \exp(-\nu - A^*X_{t-1}))^{-1}$ and therefore the smallest eigenvalue of $\Gamma_t$ is lower bounded by $(3 + \exp(\tilde{\nu} + \rho \tilde{a}))^{-1}$.
\fpro

\subsubsection{Part 2}
\spro
In order to prove this part of the Theorem, we use of Markov's inequality and Lemma \ref{ExpMartingale} in the case of the Bernoulli autoregressive process. Define the sequence $(Y_n, n\in \mathbb{N})$ as 
$$Y_n \triangleq \frac{1}{T} \sum_{t=0}^{n-1} X_{t,m}(X_{t+1,\ell} - \mathbb{E}[X_{t+1,\ell}|X_t]).$$
Notice the following values:
\begin{align*}
Y_n - Y_{n-1} =& \frac{X_{n-1,m}}{T}\left(X_{n,\ell} - \mathbb{E}[X_{n,\ell} | X_{n-1}]\right)\\
M_n^k =& \sum_{i=1}^n \mathbb{E}\left[ \left( \frac{X_{i-1,m}}{T} (X_{i,\ell} - \mathbb{E}[X_{i,\ell} | X_{i-1}])\right)^k | X_1,\ldots,X_{i-1}\right].
\end{align*}
The first value shows that $\mathbb{E}[Y_n - Y_{n-1}|X_1,\ldots,X_{n-1}] = 0$ and therefore $Y_n$ (and the negative of the sequence, $-Y_n$) is a martingale. Additionally, we know $|Y_n - Y_{n-1}| \leq \frac{1}{T} \triangleq B$ and 
\begin{align*}
M_n^2 =& \sum_{i=1}^n \mathbb{E} \left[ \left( \frac{X_{i-1,m}}{T} (X_{i,\ell} - \mathbb{E}[X_{i,\ell}|X_{i-1}])\right)^2 | X_1,\ldots,X_{i-1}\right]\\
=& \frac{1}{T^2} \sum_{i=1}^n X_{i-1,m}^2 \mathbb{E}[(X_{i,\ell} - \mathbb{E}[X_{i,\ell}|X_{i-1}])^2|X_{i-1}]
\leq  \frac{n}{4T^2} \triangleq \widehat{M}_n^2
\end{align*}
where the last step follows because Bernoulli random variables are bounded by one, and the variance is bounded by $1/4$. We also need to bound $M_n^k$ as follows:
\begin{align*}
M_n^k =& \sum_{i=1}^n \mathbb{E} \left[\left(\frac{X_{i-1,m}}{T^2} (X_{i,\ell} - \mathbb{E}[X_{i,\ell}|X_{i-1}] )\right)^k|X_1,\ldots,X_{i-1}\right]\\
= &\sum_{i=1}^n \mathbb{E} \left[\left(\frac{X_{i-1,m}}{T^2} (X_{i,\ell} - \mathbb{E}[X_{i,\ell}|X_{i-1}] )\right)^{2}\left(\frac{X_{i-1,m}}{T^2} (X_{i,\ell} - \mathbb{E}[X_{i,\ell}|X_{i-1}] )\right)^{k-2}|X_{i-1}\right]\\
\leq & B^{k-2} M_n^2 
\end{align*}
We use these values to get a bound on the summation term used in Lemma $\ref{ExpMartingale}$. 
\begin{align*}
D_n \triangleq& \sum_{k\geq 2} \frac{\eta^k}{k!}M_n^k \leq \sum_{k\geq 2} \frac{\eta^k B^{k-2} M_n^2}{k!} \leq \frac{\widehat{M}_n^2}{B^2} \sum_{k\geq 2} \frac{(\eta B)^k}{k!}\triangleq \widehat{D}_n\\
\widetilde{D}_n \triangleq& \sum_{k\geq 2} \frac{\eta^k}{k!} (-1)^k M_n^k \leq \widehat{D}_n.
\end{align*}
In the above $\widetilde{D}_n$ corresponds to the sum corresponding to the negative sequence $-Y_0, -Y_1,\ldots$ which we also need to obtain the desired bound. Now we use a variant of Markov's inequality to get a bound on the desired quantity.
\begin{align*}
\Prob(|Y_n| \geq y) = &\Prob(Y_n \geq y) + \Prob(-Y_n \geq  y) \leq \mathbb{E}[e^{\eta Y_n}] e^{-\eta y} + \mathbb{E}[e^{\eta (-Y_n)}] e^{-\eta y}\\
= &\mathbb{E}[e^{\eta Y_n - D_n + D_n}] e^{-\eta y} + \mathbb{E}[e^{\eta (-Y_n) - \widetilde{D}_n + \widetilde{D}_n}] e^{-\eta y}\\
\leq & \mathbb{E}[e^{\eta Y_n - D_n}]e^{\widehat{D}_n - \eta y} + \mathbb{E}[e^{\eta(-Y_n) - \widetilde{D}_n}] e^{\widehat{D}_n - \eta y} \leq 2 e^{\widehat{D}_n - \eta y}.
\end{align*}
The final inequality comes from the use of Lemma \ref{ExpMartingale}, which states that the given terms are supermartingales with initial term equal to 1, so the entire expectation is less than or equal to 1. The final step of the proof is to find the optimal value of $\eta$ to minimize this upper bound.
\begin{align*}
\Prob(|Y_n| \geq y)  \leq 2 \exp(\widehat{D}_n - \eta y) = 2 \exp \left(\frac{\widehat{M}_n^2}{B^2}\left(e^{\eta B} - 1 - \eta B \right)-\eta y\right)
\end{align*}
Setting $\eta = \frac{1}{B} \log \left(\frac{yB}{\widehat{M}^2_n } + 1\right)$ yields the lowest such bound, giving
\begin{align*}
\Prob(|Y_n| \geq y) \leq& 2 \exp\left(\frac{\widehat{M}_n^2}{B^2}\left(\frac{yB}{\widehat{M}_n^2} - \log\left(\frac{yB}{\widehat{M}_n^2} + 1\right) \right) - \frac{y}{B} \log\left(\frac{yB}{\widehat{M}_n^2} + 1\right) \right)\\
= & 2 \exp\left(-\frac{\widehat{M}_n^2}{B^2} H\left(\frac{yB}{\widehat{M}_n^2}\right) \right)
\end{align*}
where $H(x) =  (1 + x) \log(1+x) - x$. We use the fact that $H(x) \geq \frac{3 x^2}{2(x+3)} $ for $x\geq 0$ to further simplify the bound.
$$\Prob(|Y_n| \geq y ) \leq 2 \exp\left(\frac{-3y^2}{2yB + 6 \widehat{M}_n^2} \right) = 2 \exp\left(-\frac{6 y^2 T^2}{4 y T + 3 n} \right)$$
To complete the proof, we set $n=T$ and take a union bound over all indices because $Y_T$ considered specific indices $m$ and $\ell$, which gives the bound
\begin{align*}
\Prob\Big(\max_{1 \leq i,j \leq M}& \frac{1}{T} \left| \sum_{t=0}^{T-1} X_{t-1,i} (X_{t,j} - \mathbb{E}[X_{t,j}|X_{t-1}]) \right|\geq 3\frac{\log(MT)}{\sqrt{T}}\Big) \\
\leq &\exp\left(\log(2M^2) - \frac{54\log(MT)}{12/\sqrt{T}+3}\right) \\
\leq & \frac{1}{MT}.
\end{align*}
Here we have additionally assumed that $T\geq 2$ and that $\log(MT) \geq 1$. 
\fpro

\subsection{Proof of Theorem \ref{ThmPoiss}}
\subsubsection{Part 1}
\spro
We start with the following observation: 
 \begin{align*}
\mathbb{E}[X_{\mathcal{T}_t}X_{\mathcal{T}_t}^\top|X_{\mathcal{T}_{t-1}}] = \mathbb{E}[X_{\mathcal{T}_t}|X_{\mathcal{T}_{t-1}}]\mathbb{E}[X_{\mathcal{T}_t}|X_{\mathcal{T}_{t-1}}]^\top + {\rm{Diag}}({\rm{Var}}(X_{\mathcal{T}_t}|X_{\mathcal{T}_{t-1}}))
\end{align*}
Thus $\Gamma_t$ has two parts, one is the outer product of a vector with itself, and the second is a diagonal matrix. Therefore, the smallest eigenvalue will be lower bounded by the smallest element of the diagonal matrix. In order to lower bound this variance, we must consider the two cases, one where $\mathcal{T}_{t-1} = \mathcal{T}_t - 1$ where the previous term in the sequence $\mathcal{T}$ is the previous term in the overall sequence, and the other case where $\mathcal{T}_{t-1} < \mathcal{T}_t - 1$ where the previous term is not in the sequence $\mathcal{T}$. The variance of $X_{\mathcal{T}}$ can be characterized based on these two possible situations:
 \begin{align*}
 {\rm{Var}}(X_{\mathcal{T}_{t,i}}|X_{\mathcal{T}_{t-1}}) = p {\rm{Var}}(X_{\mathcal{T}_{t,i}} | X_{\mathcal{T}_{t-1}}, \mathcal{T}_{t-1} = \mathcal{T}_t - 1) + (1-p){\rm{Var}}(X_{\mathcal{T}_{t,i}} | X_{\mathcal{T}_{t-1}}, \mathcal{T}_{t-1} < \mathcal{T}_t - 1)
 \end{align*}
 where $p$ is the probability that $\mathcal{T}_{t-1} = \mathcal{T}_t - 1$. Because variances are lower bounded by 0, we can lower bound this entire term by the first part of the sum, where $\mathcal{T}_{t-1} = \mathcal{T}_t - 1$. For this term, we know that $X_{\mathcal{T}_{t}}$ is drawn from a Poisson distribution, with the added information that each element is bounded above by $U$ because it is an element of the sequence $X_{\mathcal{T}_1}, X_{\mathcal{T}_2}, \ldots$. Thus using Lemma \ref{LemBoundPoissVar} we know that the variance of each value is lower bounded by $\frac{4}{5}\exp(\nu_i + a^{*\top}_i X_{\mathcal{T}_{t-1}})$ which can in turn be lower bounded by $\exp(\nu_{\min} + \rho a_{\min} U)$. 
Finally, since there are at least $\xi T$ elements of $1, 2, \ldots, T$ which are in the bounded set of observations, then the worst case distribution of the observations with elements greater than $U$ is that they are never consecutive. This maximizes the number of times there is a break in the sequence $\mathcal{T}_1, \mathcal{T}_2, \ldots $, which means there would be a total of $T - \xi T$ times when there was a break. Thus the probability that consecutive elements are in the set is at least $\xi$, meaning that the minimum eigenvalue of $\mathbb{E}[X_{\mathcal{T}_t}X_{\mathcal{T}_t}^\top|X_{\mathcal{T}_{t-1}}]$ is lower bounded by $\frac{4 \xi}{5} \exp(\nu_{\min} + \rho a_{\min}U)$.
 
\fpro

\subsubsection{Part 2}
\spro
To prove this part of the Theorem, we use of Markov's inequality and Lemma \ref{ExpMartingale} as they pertain specifically to our problem. Define the sequence $(Y_n, n\in \mathbb{N})$ as 
$$Y_n \triangleq \frac{1}{T} \sum_{t=0}^{n-1} X_{t,m}(X_{t+1,\ell} - \mathbb{E}[X_{t+1,\ell}|X_t]).$$
Notice the following values:
\begin{align*}
Y_n - Y_{n-1} =& \frac{X_{n-1,m}}{T}\left(X_{n,\ell} - \mathbb{E}[X_{n,\ell} | X_{n-1}]\right)\\
M_n^k =& \sum_{i=1}^n \mathbb{E}\left[ \left( \frac{X_{i-1,m}}{T} (X_{i,\ell} - \mathbb{E}[X_{i,\ell} | X_{i-1}])\right)^k | X_1,\ldots,X_{i-1}\right].
\end{align*}
The first value shows that $\mathbb{E}[Y_n - Y_{n-1}|X_1,\ldots,X_{n-1}] = 0$ and therefore $Y_n$ (and the negative of the sequence, $-Y_n$) is a martingale. Additionally, we have assumed that $|X_{m,i}| \leq C \log M T$ for $1 \leq m \leq M$ and $1 \leq i \leq T$, so it is true that $|Y_n - Y_{n-1}| \leq \frac{C^2 \log^2(MT)}{T} \triangleq B$. Additionally:
\begin{align*}
M_n^2 =& \sum_{i=1}^n \mathbb{E} \left[ \left( \frac{X_{i-1,m}}{T} (X_{i,\ell} - \mathbb{E}[X_{i,\ell}|X_{i-1}])\right)^2 | X_1,\ldots,X_{i-1}\right]\\
=& \frac{1}{T^2} \sum_{i=1}^n X_{i-1,m}^2 \mathbb{E}[(X_{i,\ell} - \mathbb{E}[X_{i,\ell}|X_{i-1}])^2|X_{i-1}]\\
=& \frac{1}{T^2} \sum_{i=1}^n X_{i-1,m}^2 \exp(\nu_\ell + a_\ell^{*\T}X_{i-1})
\leq \frac{n C^2 \log^2(MT) e^{\nu_{\max}}}{T^2} \triangleq \widehat{M}_n^2
\end{align*}
where the last step follows because $X_{\ell,i}|X_{i-1} \sim \Poisson ( \exp(\nu_\ell + a_\ell^{*\T}X_{i-1}))$ and the mean and variance of a Poisson random variable are equal. The final line uses the fact that $X_t$ is bounded. We will also need to bound $M_n^k$ as follows:
\begin{align*}
M_n^k =& \sum_{i=1}^n \mathbb{E} \left[\left(\frac{X_{i-1,m}}{T^2} (X_{i,\ell} - \mathbb{E}[X_{i,\ell}|X_{i-1}] )\right)^k|X_1,\ldots,X_{i-1}\right]\\
= &\sum_{i=1}^n \mathbb{E} \left[\left(\frac{X_{i-1,m}}{T^2} (X_{i,\ell} - \mathbb{E}[X_{i,\ell}|X_{i-1}] )\right)^{2}\left(\frac{X_{i-1,m}}{T^2} (X_{i,\ell} - \mathbb{E}[X_{i,\ell}|X_{i-1}] )\right)^{k-2}|X_{i-1}\right]\\
\leq & B^{k-2} M_n^2 
\end{align*}
We need to use these values to get a bound on the summation term used in Lemma $\ref{ExpMartingale}$. 
\begin{align*}
D_n \triangleq& \sum_{k\geq 2} \frac{\eta^k}{k!}M_n^k \leq \sum_{k\geq 2} \frac{\eta^k B^{k-2} M_n^2}{k!} \leq \frac{\widehat{M}_n^2}{B^2} \sum_{k\geq 2} \frac{(\eta B)^k}{k!}\triangleq \widehat{D}_n\\
\widetilde{D}_n \triangleq& \sum_{k\geq 2} \frac{\eta^k}{k!} (-1)^k M_n^k \leq \widehat{D}_n
\end{align*}
In the above $\widetilde{D}_n$ corresponds to the sum corresponding to the negative sequence $-Y_0, -Y_1,\ldots$ which we will also need to obtain the desired bound. Now we are able to use a variant of Markov's inequality to get a bound on the desired quantity.
\begin{align*}
\Prob(|Y_n| \geq y) = &\Prob(Y_n \geq y) + \Prob(-Y_n \geq  y) \leq \mathbb{E}[e^{\eta Y_n}] e^{-\eta y} + \mathbb{E}[e^{\eta (-Y_n)}] e^{-\eta y}\\
= &\mathbb{E}[e^{\eta Y_n - D_n + D_n}] e^{-\eta y} + \mathbb{E}[e^{\eta (-Y_n) - \widetilde{D}_n + \widetilde{D}_n}] e^{-\eta y}\\
\leq & \mathbb{E}[e^{\eta Y_n - D_n}]e^{\widehat{D}_n - \eta y} + \mathbb{E}[e^{\eta(-Y_n) - \widetilde{D}_n}] e^{\widehat{D}_n - \eta y} \leq 2 e^{\widehat{D}_n - \eta y}
\end{align*}
The final inequality comes from the use of Lemma \ref{ExpMartingale}, which states that the given terms are supermartingales with initial term equal to 1, so the entire expectation is less than or equal to 1. The final step of the proof is to find the optimal value of $\eta$ to minimize this upper bound.
\begin{align*}
\Prob(|Y_n| \geq y)  \leq 2 \exp(\widehat{D}_n - \eta y) = 2 \exp \left(\frac{\widehat{M}_n^2}{B^2}\left(e^{\eta B} - 1 - \eta B \right)-\eta y\right)
\end{align*}
Setting $\eta = \frac{1}{B} \log \left(\frac{yB}{\widehat{M}^2_n } + 1\right)$ yields the lowest such bound, giving
\begin{align*}
\Prob(|Y_n| \geq y) \leq& 2 \exp\left(\frac{\widehat{M}_n^2}{B^2}\left(\frac{yB}{\widehat{M}_n^2} - \log\left(\frac{yB}{\widehat{M}_n^2} + 1\right) \right) - \frac{y}{B} \log\left(\frac{yB}{\widehat{M}_n^2} + 1\right) \right)\\
= & 2 \exp\left(-\frac{\widehat{M}_n^2}{B^2} H\left(\frac{yB}{\widehat{M}_n^2}\right) \right)
\end{align*}
where $H(x) =  (1 + x) \log(1+x) - x$. We can use the fact that $H(x) \geq \frac{3 x^2}{2(x+3)} $ for $x\geq 0$ to further simplify the bound.
$$\Prob(|Y_n| \geq y ) \leq 2 \exp\left(\frac{-3y^2}{2yB + 6 \widehat{M}_n^2} \right) = 2 \exp\left(-\frac{3 y^2 T^2}{2C^2(Ty+3ne^{\nu_{\max}})\log^2(MT)} \right)$$
To prove the proof, we set $n=T$ and take a union bound over all indices because $Y_T$ considered specific indices $m$ and $\ell$, which gives the bound
\begin{align*}
\Prob\Big(\max_{i,j}& \frac{1}{T} \left| \sum_{t=0}^{T-1} X_{t-1,i} (X_{t,j} - \mathbb{E}[X_{t,j}|X_{t-1}]) \right|\geq 4 C^2 e^{\nu_{\max}} \frac{\log^3(MT)}{\sqrt{T}}\Big) \\
\leq &\exp\left(\log(2M^2) - \frac{48 C^4 \exp^{2\nu_{\max}} \log^4(MT)}{8 C^4 e^{\nu_{\max}} \log^3(MT)/\sqrt{T} + 6C_1^2 e^\nu_{\max}}\right) \\
\leq  &\exp\left(2\log(MT) - \frac{24C^2e^\nu_{\max}\log(MT)}{4C^2/\sqrt{T} + 3} \right)\\
\leq & \exp\left(-c \log(MT)\right)
\end{align*}
where $c=  \frac{24C^2e^{\nu_{\max}}-8C^2-6}{4C^2+3}$ which is positive for sufficiently large $C$. Here we have additionally assumed that $T\geq 2$ and that $\log(MT) \geq 1$. 
\fpro

\section{Discussion}
\label{sec:discussion}

Corollaries \ref{CorrBern} and \ref{CorrPoiss} provide several
important facts about the inference process. 
Primarily, if $\rho$ is fixed as a constant for
increasing $M$ (suggesting that the maximum degree of a node does not
increase with the number of nodes in a network), then the error
scales inversely with $T$, linearly with the sparsity level $s$ and
only logarithmically with the dimension $M$ in order to estimate $M^2$
parameters. These parameters will dictate how much data needs to be
collected to achieve a desired accuracy level. This rate illustrates the
idea that doing inference in sparse settings can greatly reduce the
needed amount of sensing time, especially when $s\ll M^2$. Another
quantity to notice is that we require
$T \geq \omega^{-4} \rho^3 \log(M)$. If $\rho$ is fixed as a constant for
increasing $M$, this tells us that $T$ needs to be on the order of
$ \log(M)$,
 which is significantly less than the total $M^2$
parameters which are being estimated, and therefore including the
sparsity assumption has lead to a significant gain. One final
observation from the risk bound is that it provides guidance in the
setting of the regularization parameter. We see that we would like to
set $\lambda$ generally as small as possible, since the error scales
approximately like $\lambda^2$, but we also require $\lambda$ at least
as large as $\tilde{O}(T^{-1/2})$ for the bounds to hold. The balance
between setting $\lambda$ small enough to have low error, while
maintaining that it's large enough is an equivalent argument to
needing to set $\lambda$ large enough for it to take effect, but not
too large to cause over smoothing.

\subsection{Dense rows of $A^*$}
The exponential scaling in Corollaries \ref{CorrBern} and \ref{CorrPoiss} with the maximum number
of non-zeros in a row, $\rho$, at first seems unsatisfying. However,
we can imagine a worst-case scenario where a large $\rho$ relative to
$s$ and $M$ would actually lead to very poor estimation. Consider the
case of a large star-shaped network, where every node in the network
influences and is influenced by a single node, and there are no other
edges in the network. This would correspond to a matrix with a single,
dense row and corresponding column. Therefore, we would have
$\rho = M$ and $s = 2M - 1$.  In the Poisson setting, this network would
have $M-1$ independently and identically distributed
Poisson random variables at every time with mean $\nu$, but the
central node of the network would be constantly inhibited, almost
completely. In a large network, it would be very difficult to know if
this inhibition was coming from a few strong connections or from the
cumulative effect of all the inhibitions. Additionally, since the
central node would almost never have a positive count, it would also
be difficult to learn about the influence that node has on the rest of
the network. Because of networks like this, it is important that not
only is the overall network sparse, but each row also needs to be
sparse. This requirement might seem restrictive, but it has been shown
in many real world networks that the degree of a node in the network
follows a power-law which is independent of the overall size of the
network \cite{BarAlb99}, and $\rho$ would grow slowly with growing
$M$.

\subsection{Bounded observations and higher-order autoregressive processes}
Recall that Assumption~\ref{AsLearnable} ensures that most
observations are bounded. Bounded observations are important to our
analysis because we use martingale concentration inequalities
\cite{HoudreReynaudBouret03} which depend on bounded conditional means
and conditional variances, the latter condition being equivalent to
$Z$ being strongly convex. Since the conditional means and variances
are data-dependent, bounded data (at least with high probability) is a
sufficient condition for bounded conditional means and conditional
variances.  In some settings (\eg Bernoulli), bounded observations are
natural and in Assumption~\ref{AsLearnable} $\xi = 1$.  In other
settings (\eg Poisson) there is no constant $U$ independent of $T$ that
is an upper bound for {\em all} observations with high
probability. Furthermore, if we allow $U$ to increase with $T$ in
violation of Assumption~\ref{AsLearnable}, we derive a bound on
$\|\widehat{A}-A^*\|_F^2$ that {\em increases} polynomially with
$T$. To avoid this and get the far better bound in
Theorem~\ref{ThmMain}, our proof focuses on characterizing the error
on the set $\mathcal{T}$ defined in Assumption~\ref{AsLearnable}. 

Thus far we have focused on the case where $X_{t+1,m}|X_t \sim p(\nu +
a_m^{*\top}X_t)$, a first order autoregressive process. However, we could
imagine a simple, higher-order version where $X_{t+1,m} |
X_{t-q+1},\ldots,X_t \sim p(\nu + a_m^{*\top}\sum_{i=0}^{q-1} \alpha_{i}X_{t-i})$
for some known sequence $\alpha_i$. This process could be
reformulated as a process $X_{t+1,m} | X_{t-q+1},\ldots,X_t \sim p(\nu +
a_m^{*\top}\tilde{X}_t)$ where $\tilde{X}_t \triangleq \sum_{i=0}^{q-1} \alpha_{i}X_{t-1}$, and much of the same proof techniques would still hold, especially in the case of the Bernoulli autoregressive process, where $\mathcal{T}$ is easily defined. However, in the more general GLAR case finding the right analogy to $\mathcal{T}$ in the higher space is not an obvious extension. A true order-$q$ autoregressive process where $\tilde{X}_{t+1,m} | X_{t-q+1},\ldots,X_t \sim p (\nu + \sum_{i=0}^{q-1} a_{m,i}^{*\top}X_{t-i})$ could also be formulated as an order-1 process by properly stacking vectors and matrices, however, in this case proving the key lemmas and showing that the process obeys Assumption~\ref{AsLearnable} is also an open question.

\subsection{Stationarity}
As stated in the problem formulation, we restrict our attention to
bounded matrices $A^* \in [a_{\min},a_{\max}]^{M \times M}$; in the
specific context of the log-linear Poisson autoregressive model, we
use $a_{\max} = 0$, corresponding to a model that only accounts for
inhibitory interactions. One might ask whether these constraints could
be relaxed and whether the Poisson model could also account for
stimulatory interactions. 

These boundedness constraints are sufficient to
ensure that the observed process has a stationary distribution. The
stationarity of processes is heavily studied; once a process has
reached its stationary distribution, then data can be approximated as
independent samples from this distribution and temporal dependencies
can be can be ignored. While stationarity does not play an explicit
role in our analysis, we can identify several sufficient conditions to
ensure the vector GLAR processes of interest are stationary.
In particular we assume that $A^* = A^{*\T}$ which ensures
reversibility of the Markov chain described by the process defined by
$X_{t+1,m}|X_t \sim p(\nu_m + a_m^{*\top}X_t)$. We derive the stationary distribution
$\pi(x)$, and then establish bounds on the mixing time. Note that this
is a Markov chain with transition kernel:
$$ P(x,y) = \Prob(X_{t+1}=y | X_t = x) = \exp\left(\nu^\T y + y^\T A^* x - \sum_{i=m}^M Z(\nu_m + a_m^{*\T} X)\right)\prod_{m = 1}^M h(y_m) .$$
If we further assume that the entries
of $X_t$ take on values on a countable domain to ensure a
countable Markov chain, we can derive bounds on the mixing time.
\blems
\label{Lem:MixTime}
Assume $A^* = A^{*\T}$, then 
the Markov chain $X_{t+1,m} \sim p(\nu_m + a_m^{*\top}X_t)$ is a reversible Markov chain with stationary distribution:
\begin{equation*}
\pi(x) = C_{\nu,A^*} \exp\left(\nu^{\T}x + \sum_{m=1}^M Z(\nu_m + a_m^{*\T}x)\right)\prod_{m=1}^M h(x_m)
\end{equation*}
for $C_{\nu,A^*} = \int_{x_1}\int_{x_2}\ldots\int_{x_M} \exp\left(\nu^{\T}x + \sum_{m=1}^M Z(\nu_m + a_m^{*\T}x)\right)\prod_{m=1}^M h(x_m) dx_{M} \ldots dx_2 dx_1$. Further, if $X_t \in \mathbb{Z}_+^M, a_{\max} = 0$ and $Z(\cdot)$ is an increasing function, then for any $y \in \mathbb{Z}_+^M$, if $\nu_m \leq \nu_{\max} < \infty$ for all $1 \leq m \leq M$ and $a_{\min} \leq 0$ we have that
\begin{equation*}
\|P^t(y,.) - \pi(.) \|_{TV} \leq \left(1 - {h(0)^{-2M}e^{-2 M Z(\nu_{\max})}}\right)^t.
\end{equation*}
\elems

Notice that for large $M$, the chain will mix very slowly, and additionally this bound has no dependence on the sparsity of the true matrix $A^*$. Conversely,
our results require $T$ to be greater than a value that scales roughly like $\rho^3 \log(M)$, which has a much milder dependence on $M$, and varies based
on the sparsity of $A^*$. What we can conclude from these observations is that while the RMLE needs a certain amount of observations
to yield good results, we do not necessarily need enough data to reach
the stationary distribution. Additionally, under conditions where mixing time
guarantees are not given (i.e. non-symmetric $A^*$, uncountable domain), we still have guarantees on the performance of the RMLE.

\section{Conclusions}
Instances of the generalized linear autoregressive process has been used
successfully in many settings to learn network structure. 
However, this model is often used without rigorous non-asymptotic
guarantees of accuracy. In this paper we have shown important
properties of the Regularized Maximum Likelihood Estimator of the GLAR
process under a sparsity assumption. We have proven bounds on the
error of the estimator as a function of sparsity, maximum degree of a
node, ambient dimension and time, and shown how these bounds look for
the specific examples of the Bernoulli and Poisson autoregressive
proceses. In order to prove this risk bound, we have incorporated many
recently developed tools of statistical learning, including
concentration bounds for dependent random variables. Our results show
that by incorporating sparsity the amount of data needed is on the
order of $\rho^3 \log (M)$ for bounded degree networks, which is a
significant gain compared to the $M^2$ parameters being
estimated. 

While this paper has focused on generalized linear models, we believe that
the extension of these ideas to other models is
possible. Specifically, for modeling firing rates of neurons in the brain, we are interested in settings in which we
observe
$$X_{t+1,m}|X_t \sim \mbox{Poisson}(g(a_m^{*\top} X_t + \nu))$$
and exploring possible functions $g$ beyond the exponential function
considered here. Such analysis would allow our results to apply to
stimulatory effects in addition to inhibitory effects, but key
challenges include ensuring that the process is stable and, with high
probability, bounded. Another direction would be settings where the counts are drawn from more complicated higher-order or autoregressive moving average (ARMA) models which would better model real-world point processes.

\section{Appendix}
\label{ref:Appendix}
\subsection{Supplementary Lemmas}

First we present supplementary Lemmas which we use throughout the proofs of the main Theorems. 

\blems
\label{LemBoundPoissVar}
Let $X$ be a Poisson random variable, with the following probability density function:
$$ p(X = k | \lambda) = \frac{\lambda ^k e ^{-k}}{k!}$$
and let $X'$ be a random variable defined by the following pdf:
$$q(k|\lambda)=\begin{cases} \frac{c}{k!}\lambda^k e^{-\lambda} \text{ if } k\leq U \\ 0 \text{ otherwise}  \end{cases}$$
where $c=\frac{1}{1-\mathbb{P}(X>U)}>1.$ Roughly speaking, $X'$ is generated by taking a Poisson pdf, and removing the tail probability, and scaling the remaining density so that it is a valid pdf. For this random variable, assuming $U \geq \max(6, 1.5 e \lambda, \lambda + 5)$ then
$${\rm{Var}}(X') \geq \frac{4}{5} {\rm{Var}}(X) = \frac{4 \lambda}{5}$$ 

\elems
\spro
Define the error terms  $\epsilon_1 \triangleq \mathbb{E}[X]^2-\mathbb{E}[X']^2$ and $\epsilon_2 \triangleq \mathbb{E}[X^2]-\mathbb{E}[X'^2]$. We know 
\begin{align}
\label{eq:boundedDistVar}
{\rm{Var}}(X')=\mathbb{E}[X'^2]-\mathbb{E}[X']^2=(\mathbb{E}[X^2]-\epsilon_2)-(\mathbb{E}[X]^2 -\epsilon_1) \nonumber\\
\geq \underbrace{(\mathbb{E}[X^2]-\mathbb{E}[X]^2)}_{{\rm{Var}}(X)}-(|\epsilon_1|+|\epsilon_2|)= \lambda-(|\epsilon_1|+|\epsilon_2|)
\end{align}  
Our strategy will be to show $\epsilon_1,\epsilon_2$ are small relative to $\lambda$, which will tell us ${\rm{Var}}(X') \approx {\rm{Var}}(X)=\lambda$.  Intuitively, the error terms should be small relative to $\lambda$ because $X'$ differs from $X$ only by cutting off the extreme edge of the pdf, given the assumptions on the size of $U$ relative to $\lambda$. 

First, we bound $\epsilon_1$.  We have 
$$\epsilon_1=\mathbb{E}[X]^2-\mathbb{E}[X']^2= (\mathbb{E}[X]+\mathbb{E}[X'])(\mathbb{E}[X]-\mathbb{E}[X'])$$  
Since $\mathbb{E}[X'] \leq \mathbb{E}[X]$, the first term is bounded by $2\mathbb{E}[X]=2\lambda$. To bound the second term, we note that the pdf for $X'$ is given explicitly as 
$$q(k|\lambda)=\begin{cases} \frac{c}{k!}\lambda^k e^{-\lambda} \text{ if } k\leq U \\ 0 \text{ otherwise}  \end{cases}$$ 
where $c=\frac{1}{1-\mathbb{P}(X>U)}>1.$ And therefore 
\begin{equation*}
\mathbb{E}[X']=c \sum_{k=1}^U \frac{\lambda^k e^{-\lambda}}{(k-1)!} \geq  \sum_{k=1}^U \frac{\lambda^k e^{-\lambda}}{(k-1)!}
\end{equation*} 
Using this fact to bound $\mathbb{E}[X]-\mathbb{E}[X']$ gives us
\begin{equation*}
\mathbb{E}[X]-\mathbb{E}[X'] \leq \mathbb{E}[X]-\sum_{k=1}^U \frac{\lambda^k e^{-\lambda}}{(k-1)!}=\sum_{k=U+1}^\infty \frac{\lambda^k e^{-\lambda}}{(k-1)!}=\frac{\lambda}{e^\lambda} \sum_{k=U}^\infty \frac{\lambda^k}{k!}
\end{equation*} 
Note $\sum_{k=U}^\infty \frac{\lambda^k}{k!}$ is the remainder term of the degree $U-1$ Taylor Polynomial for $e^\lambda$.  We can bound this using Taylor's Remainder theorem: 
$$\sum_{k=U}^\infty \frac{\lambda^k}{k!} \leq \frac{e^\lambda \lambda^{U}}{U!}$$ 
and so 
\begin{equation*}
\mathbb{E}[X]-\mathbb{E}[X'] \leq \lambda \frac{\lambda^U}{U!} \leq \frac{\lambda}{1.5^U} \frac{(\frac{U}{e})^U}{U!}
\end{equation*}
where the second inequality comes from the assumption that $U \geq 1.5e \lambda$.
Here, the second fraction is small by Sterling's approximation formula.  Formally, Sterling tells us 
$$\frac{(\frac{U}{e})^U}{U!} \leq \frac{1}{\sqrt{2\pi U}}$$ 
and therefore 
\begin{equation*} 
\mathbb{E}[X]-\mathbb{E}[X'] \leq \frac{\lambda}{1.5^U \sqrt{2\pi U}}.
\end{equation*}  
Combining the two terms tells us 
$$|\epsilon_1| \leq  2\lambda \frac{\lambda}{1.5^U \sqrt{2\pi U}} \leq \frac{\lambda}{10}$$ 
since $U \geq 6$. 

Next we bound $\epsilon_2=\mathbb{E}[X^2]-\mathbb{E}[X'^2]$.  We have 
$$\mathbb{E}[X'^2]=c \sum_{k=1}^U \frac{k \lambda^k e^{-\lambda}}{(k-1)!} \geq \sum_{k=1}^U \frac{k \lambda^k e^{-\lambda}}{(k-1)!}$$ 
and therefore
$$\epsilon_2 \leq \mathbb{E}[X^2]-\sum_{k=1}^U \frac{k \lambda^k e^{-\lambda}}{(k-1)!}=\sum_{k=U+1}^\infty \frac{k \lambda^k e^{-\lambda}}{(k-1)!}\leq \frac{(U+1)\lambda^2}{Ue^\lambda}\sum_{k=U-1}^\infty \frac{ \lambda^k }{k!}$$ 
where the last inequality is due to the fact that $\frac{k}{k-1} \leq \frac{U+1}{U}$ for all $k \geq U + 1$. Here $\sum_{k=U-1}^\infty \frac{ \lambda^k }{k!}$ is the remainder term for the degree $U-2$ Taylor Polynomial approximation to $e^\lambda$.  By the Taylor's remainder formula, we can bound this by 
$$\frac{e^\lambda \lambda^{U-1}}{(U-1)!}$$ and so $$|\epsilon_1| \leq \lambda (U+1) \frac{\lambda^U}{U!}$$ 
and since $\lambda \leq \frac{U}{1.5e}$, it follows from Sterling's approximation that 
$$|\epsilon_1| \leq \lambda  \frac{U+1}{1.5^{U} \sqrt{2\pi U}} \leq \frac{\lambda}{10}$$ since $U\geq 6$.  
 
Putting the bounds for $\epsilon_1$ and $\epsilon_2$ back into Equation \ref{eq:boundedDistVar} to get the final form of the Lemma
$$ {\rm{Var}}(X') \geq \frac{4}{5}{\rm{Var}}(X) = \frac{4}{5}\lambda$$
\fpro

We next present a one-sided concentration bound for Poisson random variables due to Bobkov and Ledoux~\cite{BobkovLedoux}:
random variables.
\blems[Proposition 10 in~\cite{BobkovLedoux}]
\label{LemConc}
If $X \sim \mbox{Poisson}(\lambda)$:
\begin{equation*}
\mathbb{P}\big(X - \lambda > t  \big) \leq \exp(-\frac{t}{4}\log(1+\frac{t}{2 \lambda})).
\end{equation*}
\elems

\blems[Lemma 3.3 in~\cite{HoudreReynaudBouret03}]
\label{ExpMartingale}
Let $(Y_n, n\in \mathbb{N})$ be a martingale. For all $k\geq 2$, let 
$$M_n^k \triangleq \sum_{i=1}^n \mathbb{E}[(Y_i - Y_{i-1})^k| \mathcal{F}_{i-1}].$$

Then for all integers $n\geq 1$ and for all $\eta$ such that for all $i \leq n$, $\mathbb{E}[\exp(|\eta (Y_i - Y_{i-1})|)] \leq \infty$,
$$ \varepsilon_n \triangleq \exp \left( \eta Y_n - \sum_{k\geq 2} \frac{\eta^k}{k!} M_n^k \right)$$
is a super-martingale. Additionally, if $Y_0 = 0$, then $\mathbb{E}[\varepsilon_n] \leq 1$.
\elems

\blems
\label{LemRadComp}
Let $(\epsilon_t)_{t=0}^T$ be i.i.d.\ Rademacher random variables (i.e. $\mathbb{P}(\epsilon_t = +1) = \mathbb{P}(\epsilon_t = -1) = -1 $ and $\{X_t\}_{t=0}^T$ are a sequence of random variables, where $X_t \in [0,U]^M$, $X_t(\epsilon_1, \epsilon_2,\ldots,\epsilon_{t-1})$ is a function of $(\epsilon_1,\epsilon_2,\ldots,\epsilon_{t-1})$. Then
\begin{equation*}
\sup_{X_1,\ldots,X_T} \|\frac{1}{T} \sum_{t=1}^T X_{t}(\epsilon_1, \epsilon_2,\ldots,\epsilon_{t-1}) \epsilon_t \|_\infty \leq 
2U\frac{\log(MT)}{\sqrt{T}},
\end{equation*}
with probability at least $1 - \frac{1}{(MT)^2}$.
\elems

\spro To prove this Lemma, we once again use Markov's
  inequality and Lemma \ref{ExpMartingale}. For a fixed
  $m \in \{1,\ldots,M\}$, define the sequence $(Y_n, n\in \mathbb{N})$
  as
$$Y_n \triangleq \frac{1}{T} \sum_{t=1}^{n} X_{t,m}\epsilon_t.$$
Notice the following values:
\begin{align*}
Y_n - Y_{n-1} =& \frac{1}{T}\epsilon_n X_{n,m} \\
M_n^k =& \sum_{t=1}^n \mathbb{E}\left[ \left( \frac{1}{T} X_{t,m} \epsilon_t \right)^k | \epsilon_1,\ldots,\epsilon_{t-1}\right].
\end{align*}
The first value shows that $\mathbb{E}[Y_n - Y_{n-1}|\epsilon_1,\ldots,\epsilon_{n-1}] = 0$ and therefore $Y_n$ (and the negative of the sequence, $-Y_n$) is a martingale. Additionally, we have assumed that $0 \leq X_{m,i} \leq U$ for $1 \leq m \leq M$ and $1 \leq i \leq T$, so it is true that $|Y_n - Y_{n-1}| \leq \frac{2U}{T} \triangleq B$. Additionally:
\begin{align*}
M_n^2 =& \sum_{t=1}^n \mathbb{E} \left[ \left( \frac{1}{T}X_{t,m} \epsilon_t \right)^2 | \epsilon_1,\ldots,\epsilon_{t-1}\right]\\
=& \frac{1}{T^2} \sum_{t=1}^n \epsilon_t^2 \mathbb{E} \left[X_{t, m}^2|\epsilon_1,\ldots,\epsilon_{t-1}\right]\\
&\leq \frac{4 n U^2}{T^2} \triangleq \widehat{M}_n^2
\end{align*}
We will also need to bound $M_n^k$ as follows:
\begin{align*}
M_n^k =& \sum_{t=1}^n \mathbb{E} \left[\left(\frac{\epsilon_t}{T^2} X_{t, m}\right)^k|\epsilon_1,\ldots,\epsilon_{t-1}\right]\\
= &\sum_{t=1}^n \mathbb{E} \left[\left(\frac{\epsilon_t}{T^2} X_{t, m}\right)^{2}\left(\frac{\epsilon_t}{T^2} X_{t, m}\right)^{k-2}|\epsilon_1,\ldots,\epsilon_{t-1}\right]\\
\leq & B^{k-2} M_n^2 
\end{align*}
We need to use these values to get a bound on the summation term used in Lemma $\ref{ExpMartingale}$. 
\begin{align*}
D_n \triangleq& \sum_{k\geq 2} \frac{\eta^k}{k!}M_n^k \leq \sum_{k\geq 2} \frac{\eta^k B^{k-2} M_n^2}{k!} \leq \frac{\widehat{M}_n^2}{B^2} \sum_{k\geq 2} \frac{(\eta B)^k}{k!}\triangleq \widehat{D}_n\\
\widetilde{D}_n \triangleq& \sum_{k\geq 2} \frac{\eta^k}{k!} (-1)^k M_n^k \leq \widehat{D}_n
\end{align*}
In the above $\widetilde{D}_n$ corresponds to the sum corresponding to the negative sequence $-Y_0, -Y_1,\ldots$ which we will also need to obtain the desired bound. Now we are able to use a variant of Markov's inequality to get a bound on the desired quantity.
\begin{align*}
\Prob(|Y_n| \geq y) = &\Prob(Y_n \geq y) + \Prob(-Y_n \geq  y) \leq \mathbb{E}[e^{\eta Y_n}] e^{-\eta y} + \mathbb{E}[e^{\eta (-Y_n)}] e^{-\eta y}\\
= &\mathbb{E}[e^{\eta Y_n - D_n + D_n}] e^{-\eta y} + \mathbb{E}[e^{\eta (-Y_n) - \widetilde{D}_n + \widetilde{D}_n}] e^{-\eta y}\\
\leq & \mathbb{E}[e^{\eta Y_n - D_n}]e^{\widehat{D}_n - \eta y} + \mathbb{E}[e^{\eta(-Y_n) - \widetilde{D}_n}] e^{\widehat{D}_n - \eta y} \leq 2 e^{\widehat{D}_n - \eta y}
\end{align*}
The final inequality comes from the use of Lemma \ref{ExpMartingale}, which states that the given terms are supermartingales with initial term equal to 1, so the entire expectation is less than or equal to 1. The final step of the proof is to find the optimal value of $\eta$ to minimize this upper bound.
\begin{align*}
\Prob(|Y_n| \geq y)  \leq 2 \exp(\widehat{D}_n - \eta y) = 2 \exp \left(\frac{\widehat{M}_n^2}{B^2}\left(e^{\eta B} - 1 - \eta B \right)-\eta y\right)
\end{align*}
Setting $\eta = \frac{1}{B} \log \left(\frac{yB}{\widehat{M}^2_n } + 1\right)$ yields the lowest such bound, giving
\begin{align*}
\Prob(|Y_n| \geq y) \leq& 2 \exp\left(\frac{\widehat{M}_n^2}{B^2}\left(\frac{yB}{\widehat{M}_n^2} - \log\left(\frac{yB}{\widehat{M}_n^2} + 1\right) \right) - \frac{y}{B} \log\left(\frac{yB}{\widehat{M}_n^2} + 1\right) \right)\\
= & 2 \exp\left(-\frac{\widehat{M}_n^2}{B^2} H\left(\frac{yB}{\widehat{M}_n^2}\right) \right)
\end{align*}
where $H(x) =  (1 + x) \log(1+x) - x$. We can use the fact that $H(x) \geq \frac{3 x^2}{2(x+3)} $ for $x\geq 0$ to further simplify the bound.
$$\Prob(|Y_n| \geq y ) \leq 2 \exp\left(\frac{-3y^2}{2yB + 6 \widehat{M}_n^2} \right) = 2 \exp\left(-\frac{3 y^2 T^2}{8(Ty+3n)} \right)$$
To complete the proof, we set $n=T$ and take a union bound over all indices because $Y_T$ considered specific indices $m$, which gives the bound
\begin{align*}
\Prob\Big(\max_{m}& \frac{1}{T} \left| \sum_{t=1}^{T} X_{t,m} \epsilon_t \right|\geq 2U \frac{\log(MT)}{\sqrt{T}}\Big) \\
\leq &\exp\left(\log(M) - \frac{12 U^2 T \log^2(MT)}{4 U^2  (\sqrt{T} \log(MT) + 3T)}\right) \\
\leq  &\exp\left(\log(MT) - \frac{3\log(MT)}{1/\sqrt{T} + 3/\log(MT)} \right)\\
\leq & \exp\left(-2 \log(MT)\right).
\end{align*}
\fpro

\subsection{Proof of Lemma \ref{LemPoissBound}}
\subsubsection{Part 1}
\spro
For all $1 \leq t \leq T$ and $1 \leq m \leq M$, $X_{t,m}|X_{t-1}$ is drawn from a Poisson distribution with mean $e^{\nu_m + a_m^{*\T}X_{t-1}}$ for some $a_m^* \in [a_{\min}, 0]^{M}$. Because of the range of values $a_m^*$ can take, we know that $e^{\nu_m + a_m^{*\T}X_{t-1}} \leq e^{\nu_{\max}}$ where $\nu_m \leq \nu_{\max}$ for some $\nu_{\max} < \infty$ for all $m$. Therefore, we know that
$$
\mathbb{P}(X_{t,m} \geq \eta + e^{\nu_{\max}}|X_{t-1}) \leq \mathbb{P}(Y \geq \eta + e^{\nu_{\max}})
$$
where $Y$ is a Poisson random variable with mean $e^{\nu_{\max}}$. To bound this quantity we use the result of Lemma~\ref{LemConc},
\begin{equation*}
\mathbb{P}(Y > \eta+e^{\nu_{\max}}) \leq \exp\left(-\frac{\eta}{4} \log\left(1+\frac{\eta}{2 e^{\nu_{\max}}}\right) \right).
\end{equation*}
Setting $\eta = C\log MT - e^{\nu_{\max}}$,
\begin{eqnarray*}
\mathbb{P}(Y > C\log MT) &\leq& \exp\left(-\frac{C\log MT - e^{\nu_{\max}}}{4} \log\left(1 + \frac{C\log MT-e^{\nu_{\max}}}{2e^{\nu_{\max}}}\right)\right) \\
&\leq& \exp\left(-\frac{C\log MT - e^{\nu_{\max}}}{4}\right).
\end{eqnarray*}
Here, we have assumed that $C \geq e^{\nu_{\max}}(2e - 1)$ and $\log MT \geq 1$. This upper bound is not dependent on the value of $X_{t-1}$, so this quantity is also an upper bound for the unconditional probability of $X_{t,m} \geq C \log MT$. Using this for a single index $t,m$ of our data $X$, and taking a union bound over all possible indices $1\leq m \leq M, 1\leq t \leq T$ gives
\begin{equation}
\Prob\left(\max_{1\leq m \leq M, 1 \leq t \leq T} X_{t,m} > C \log M T\right ) \leq \exp\left(\log MT -\frac{C\log MT - e^{\nu_{\max}}}{4}\right) \leq \exp(-c \log MT)
\end{equation} 
for $c \leq \frac{C-e^{\nu_{\max}}}{4} -1.$ Thus if $C > \max(e^{\nu_{\max}}(2e-1), 4 + e^{\nu_{\max}})$, then $c > 0$, and the bound is valid.
\fpro

\subsubsection{Part 2}
\spro
We are interested in bounding the number of observations $X_{t,m}$ for $1\leq m \leq M$ and $1 \leq t \leq T$ that are above the value $U$. Saying at least $j \triangleq \alpha MT$ observations are less than a certain value, is equivalent to saying that the $j^{th}$ smallest observation is less than that value. Therefore, 
\begin{align*}
\Prob&(j^{th} {\text{smallest observation  }} X_{t,m} > U) = \Prob\left(\sum_{t=1}^T\sum_{m=1}^M Y_{t,m} \leq j-1\right)\\
 &= \sum_{\ell = 0}^{j-1} \Prob\left(\sum_{t=1}^T\sum_{m=1}^M Y_{t,m} = \ell\right)
 \leq \sum_{\ell=0}^{j} \sum_{y \in \mathcal{Y}^\ell} \Prob(Y=y).
\end{align*}
Here we define $Y_{t,m} \triangleq \boldsymbol{1}\{X_{t,m} \leq U \}$, and $\mathcal{Y}^\ell = \{ y \in \{0,1\}^{M \times T} | \sum_{t=1}^T \sum_{m=1}^M y_{t,m} = \ell\}.$ We then condition the values of $Y_t$ on all previous values of $Y$ and then understand this as a marginal of the joint distribution over $Y_t$ and $X_{t-1}$. Below we use the notation $Y_{1:t}$ to denote all the time indices of $Y$ from 1 to $t$, and similarly for $y$.
\begin{align*}
\Prob(Y=y) = &\prod_{t=1}^T \Prob(Y_t = y_t | Y_{1:t-1} = y_{1:t-1}) \\
= & \prod_{t=1}^T \sum_{x_{t-1}} \left(\Prob(Y_t = y_t | Y_{1:t-1} = y_{1:t-1}, X_{t-1} = x_{t-1}) \Prob(X_{t-1} = x_{t-1} | Y_{1:t-1} = y_{1:t-1})\right)\\
=& \prod_{t=1}^T \sum_{x_{t-1}}\left( \left(\prod_{m=1}^M \Prob(Y_{t,m}=y_{t,m}|X_{t-1} = x_{t-1})\right)\Prob(X_{t-1} = x_{t-1} | Y_{1:t-1} = y_{1:t-1})\right)
\end{align*}
In the last line we use the fact that conditioned on $X_{t-1}$, $Y_t$ is independent across dimensions $m$, and independent of previous values $Y_{1:t-1}$. We now make the observation that $\Prob(X_{t,m} > U |X_{t-1} = x_{t-1})$ is exactly the probability that a Poisson random variable with rate $\exp(\nu_m + a_m^{*\T}x_{t-1})$ is greater than $U$, which can be upper-bounded by the probability that a Poisson random variable with rate $\exp(\nu_{\max})$ is greater than $U$ because we have assumed all values of $a_m^*$ are non-positive. Call this probability $p_{\nu_{\max}}$. Thus we have $\Prob(Y=y) \leq p_{\nu_{\max}}^{MT - \sum_{t=1}^T\sum_{m=1}^m y_{t,m}}$ and therefore,

\begin{align*}
\Prob\left(\sum_{t=1}^T\sum_{m=1}^M Y_{t,m} \leq j-1\right) \leq & \sum_{\ell=0}^{j} \binom{MT}{\ell}p_{\nu_{\max}}^{MT - \ell} = (1+p_{\nu_{\max}})^{MT} - \sum_{\ell=0}^{MT - j-1} \binom{MT}{\ell} p_{\nu_{\max}}^{\ell}\\
\leq & \binom{MT}{MT - j } (1+p_{\nu_{\max}})^{j} p_{\nu_{\max}}^{MT - j } \leq \left(\frac{MTe}{MT-j}\right)^{MT-j} (1+p_{\nu_{\max}})^{j}p_{\nu_{\max}}^{MT-j}.
\end{align*}
The second inequality is from the application of Taylor's Remainder Theorem, and the third is from the fact that $\binom{n}{k} \leq \left(\frac{ne}{k}\right)^k$. Now use the fact that $j = \alpha MT $ as stated in the Lemma, to give
\begin{align*}
\Prob\left(\sum_{t=1}^T\sum_{m=1}^M Y_{t,m} \leq j-1\right) \leq & 
\left( \frac{p_{\nu_{\max}} e}{1- \alpha} \right)^{(1-\alpha)MT}(1+p_{\nu_{\max}})^{\alpha MT} 
\leq \left[\left(\frac{p_{\nu_{\max}} e}{1-\alpha}\right)^{1-\alpha} 2^\alpha \right] ^ {MT}.
\end{align*}
By using Lemma \ref{LemConc} in a similar way as was used in the proof of Lemma \ref{LemPoissBound} part 1, $p_{\nu_{\max}}$ can be controlled by $U$ in the following way,
\begin{align*}
p_{\nu_{\max}} =& \Prob \left( X > U\right) \leq \exp \left(-\frac{U - e^{\nu_{\max}}}{4} \log \left(1 + \frac{U - e^{\nu_{\max}}}{2e^{\nu_{\max}}}\right)\right) 
\leq \exp\left(-\frac{U - e^{\nu_{\max}}}{4}\right),
\end{align*}
when $U \geq e^{\nu_{\max}} (2 e -1)$. Plugging the result back into the bound gives
\begin{align*}
\Prob\left(\sum_{t=1}^T\sum_{m=1}^M Y_{t,m} \leq j -1 \right) \leq &
\left[\left(\frac{\exp(1-(U - e^{\nu_{\max}})/4)}{1-\alpha} \right)^{1-\alpha} 2^\alpha \right]^{MT}.
\end{align*}
When $U > 4 + e^{\nu_{\max}} + \frac{4 \alpha \log(2)}{1-\alpha} - 4 \log(1-\alpha)$ and additionally greater than $e^{\nu_{\max}}(2e-1)$ the condition from above, then the probability of this event is decaying in $M$ and $T$. Therefore, for $c = -\left( 1 - \frac{U - e^{\nu_{\max}} }{4} - \log(1-\alpha) \right) (1 - \alpha) - \alpha \log(2)$, we have the inequality
$$ \mathbb{P}( {\text{at least }} \alpha MT {\text{ observations }} X_{t,m} \leq U) \geq 1 - e^{-cMT} $$
 \fpro

\subsection{Proof of Lemma \ref{Lem:MixTime}}
\spro
To prove the form of the stationary distribution we show that
\begin{equation*}
\pi(y) = \int_x{\pi(x)P(x,y)},
\end{equation*}
where 
\begin{equation*}
P(x,y) = \exp\left(\nu^\T y + y^\T A^* x - \sum_{m=1}^M Z(\nu_m+ a_m^\T x)\right)\prod_{m = 1}^M h(y_m).
\end{equation*}
Plugging in $\pi(x)$ as specified,
\begin{align*}
\int_x \pi(x) P(x,y) =&C_{\nu,A^*}\int_x \exp\left(\nu^{\T}x + \sum_{m=1}^M Z(\nu_m + a_m^{*\T}x) + \nu^\T y + y^\T A^* x - \sum_{m=1}^M Z(\nu_m+ a_m^{*\T} x)\right)\prod_{m=1}^M h(x_m) h(y_m)\\
= &C_{\nu,A^*} \exp(\nu^\T y) \prod_{m=1}^M \left(h(y_m)\right) \int_x \exp \left( \nu ^\T x + y ^\T A^* x \right) \prod_{m=1}^M h(x_m)\\
= & C_{\nu,A^*}\exp(\nu ^\T y) \prod_{m=1}^M \left(h(y_m)\right) \int_x \exp \left( \nu ^\T x + x^\T A^* y \right) \prod_{m=1}^M h(x_m)\\
= &C_{\nu,A^*} \exp(\nu ^\T y) \prod_{m=1}^M \left( h(y_m) \int_{x_m} \exp \left(\nu_m x_m + x_m a_m^{*\T} y \right) h(x_m)\right)\\
= &C_{\nu,A^*} \exp\left(\nu ^\T y + \sum_{m=1}^M Z(\nu_m + a_m^{*\T}y) \right) \prod_{m=1}^M h(y_m) = \pi(y)
\end{align*}
The second to last equality uses the definition of $Z$ as the log partition function, and the third uses the assumption that $A^* = A^{*\T}$.

To prove the upper bound on total variation distance for Markov chains on countable domains, we define two chains, one chain $Y_t$ begins at the stationary distribution and the other independent chain starts at $X_t$ begins at some arbitrary random state $x$, both with transition kernel $P$. These two chains are said to be coupled if they are run independently until the first time where the states are equal, then are equal for the rest of the trial. The notation $P^t(x,y)$ denotes the probability of transitioning from state $y$ to state $x$ in exactly $t$ steps. Theorem 5.2 of ~\cite{LevinPeresBook} asserts that:
\begin{equation*}
\|P^t(x,\cdot) - \pi(\cdot) \|_{TV} \leq \mathbb{P}_x(\tau_{couple} > t),
\end{equation*}
where $\tau_{couple} := \left\{\displaystyle\min_{t > 0} : X_t = Y_t\right\}$. Note first that $\mathbb{P}(\tau_{couple} > t) \leq\displaystyle \prod_{\tau=0}^{t}(1 - \mathbb{P}(X_\tau = Y_\tau=0))$. Since the chains are independent until $\tau_{couple}$, $\mathbb{P}(X_\tau = Y_\tau=0) = \mathbb{P}(X_\tau = 0) \mathbb{P}(Y_\tau = 0)$. Note also that:
\begin{align*}
\mathbb{P}(X_\tau = 0 | X_{\tau-1} = x) =& h(0)^M \exp\left(-\sum_{m=1}^MZ(\nu_m + a_m^{*\T}x)\right) \\ 
\geq& h(0)^M\exp\left({-\sum_{m=1}^M Z(\nu_m)}\right) \geq h(0)^M\exp( - MZ(\nu_{\max})),
\end{align*}
where the first inequality is due to the fact that $Z$ is an increasing function, and from the assumption that $A_{i,j} \geq 0$. Hence $\mathbb{P}(\tau_{couple} > t) \leq \prod_{\tau=0}^{t}(1 - h(0)^{-2M}\exp(-2MZ(\nu_{\max}))) = (1 - h(0)^{-2M}\exp(-2MZ(\nu_{\max})))^t$. 
\fpro

\subsection{Empirical processes for martingale sequences}
\label{App:Emp}

\newcommand{\bfx}{\textbf{x}}

To concretely define the martingale, let $(X_t)_{t \geq 1}$ be a sequence of random variables adapted to the filtration $(\mathcal{A}_t)_{t \geq 1}$. First we present a bounded difference inequality for martingales developed by van de Geer~\cite{VDGMartingale}.

\btheos[Theorem 2.6 in ~\cite{VDGMartingale}]
\label{ThmMcDiarmid}
Fix $T \geq 1$ and let $Z_T$ be an $\mathcal{A_T}$-measurable random variable, satisfying for each $t = 1,2,\ldots,t$, $$L_t \leq \mathbb{E}[Z_T|\mathcal{A}_t] \leq U_t,$$ almost surely where $L_t < U_t$ are constants. Define $C_T^2 = \sum_{t=1}^{T}(U_t-L_t)^2$. Then for all $a > 0$,
$$
\mathbb{P}(Z_T - \mathbb{E}[Z_T] \geq a) \leq \exp(-\frac{2 a^2}{C_T^2}).
$$
\etheos

The second important result we need is a notion of sequential Rademacher complexity for martingales that allows us to do symmetrization, an important step in empirical process theory (see e.g.~\cite{Pollard84}). To do this we use machinery developed in ~\cite{RakhlinTewari}. Recall that $(X_t)_{t \geq 1}$ is a martingale and let $\chi$ be the range of each $X_t$. Let $\mathcal{F}$ be a function class where for all $f \in \mathcal{F}$, $f : \mathcal{\chi} \rightarrow \mathbb{R}$.

To define the notion of sequential Rademacher complexity, we first let $(\epsilon_t)_{t=1}^{T}$ be a sequence of independent Rademacher random variables (i.e. $\Prob(\epsilon_t = +1) = \Prob(\epsilon_t = -1) = \frac{1}{2}$). Next we define a tree process as a function of these independent Rademacher random variables.

A $\chi$-valued tree $\bfx$ of depth $T$ is a rooted complete binary tree with nodes labelled by elements of $\chi$. We identify the tree $\bfx$ with the sequence $(\bfx_1, \bfx_2,\ldots,\bfx_T)$ of labeling functions $\bfx_t : \{\pm 1 \}^{t-1} \rightarrow \chi$ which provide the labels for each node. Here $\bfx_1 \in \chi$ is the label for the root of the tree, while $\bfx_t$ for $t > 1$ is the label of the node obtained by following the path of length $t-1$ from the root, with $+1$ indicating ``right'' and $-1$ indicating ``left.'' Based on this tree, $\bfx_t$ is a function of $(\epsilon_1, \epsilon_2,\ldots,\epsilon_{t-1})$.

Based on this, we define the sequential Rademacher complexity of a function class $\mathcal{F}$.
\bdes[Definition 3 in ~\cite{RakhlinTewari}]
The sequential Rademacher complexity of a function class $\mathcal{F}$ on a $\chi$-valued tree $\bfx$ is defined as 
$$
\mathcal{R}_T(\mathcal{F}) \triangleq \sup_{\bfx} \mathbb{E}\biggr[\sup_{f \in \mathcal{F}}\epsilon_tf(\bfx_t(\epsilon_1, \epsilon_2,\ldots,\epsilon_{t-1}) )  \biggr]
$$
\edes
where the outer supremum is taken over all $\chi$-valued trees. Importantly note that $\big(\epsilon_tf(\bfx_t(\epsilon_1, \epsilon_2,\ldots,\epsilon_{t-1}) \big)_{t\geq1}$ is a martingale. Now we are in a position to state the main result which allows us to do symmetrization for functions of martingales.

\btheos[Theorem 2 in ~\cite{RakhlinTewari}]
\label{ThmSymmetrization}
$$
\mathbb{E}\biggr[\sup_{f \in \mathcal{F}} \frac{1}{T} \sum_{t=1}^{T} {\mathbb{E}[f(X_t)| \mathcal{A}_{t-1}] - f(X_t)} \biggr] \leq 2 \mathcal{R}_T(\mathcal{F}).
$$
\etheos
For further details refer to~\cite{RakhlinTewari}.
 
\bibliographystyle{IEEEbib}
\bibliography{Biblio_PoissonAR}

\begin{thebibliography}{10}

\bibitem{brown2004multiple}
E.~N. Brown, R.~E. Kass, and P.~P. Mitra,
\newblock ``Multiple neural spike train data analysis: state-of-the-art and
  future challenges,''
\newblock {\em Nature neuroscience}, vol. 7, no. 5, pp. 456--461, 2004.

\bibitem{colemanConvexPoint}
T.~P. Coleman and S.~Sarma,
\newblock ``Using convex optimization for nonparametric statistical analysis of
  point processes,''
\newblock in {\em Proc. ISIT}, 2007.

\bibitem{SmithBrownStateSpace}
A.~C. Smith and E.~N. Brown,
\newblock ``Estimating a state-space model from point process observations,''
\newblock {\em Neural Computation}, vol. 15, pp. 965--991, 2003.

\bibitem{HinneHeskes2012}
M.~Hinne, T.~Heskes, and M.~A.~J. {van Gerven},
\newblock ``Bayesian inference of whole-brain networks,''
\newblock {\em arXiv:1202.1696 [q-bio.NC]}, 2012.

\bibitem{DingSchroeder2011}
M.~Ding, {CE} Schroeder, and X.~Wen,
\newblock ``Analyzing coherent brain networks with {G}ranger causality,''
\newblock in {\em Conf. Proc. IEEE Eng. Med. Biol. Soc.}, 2011, pp. 5916--8.

\bibitem{spikesPillow}
J.~W. Pillow, J.~Shlens, L.~Paninski, A.~Sher, A.~M. Litke, E.~J. Chichilnisky,
  and E.~P. Simoncelli,
\newblock ``Spatio-temporal correlations and visual signalling in a complete
  neuronal population,''
\newblock {\em Nature}, vol. 454, pp. 995--999, 2008.

\bibitem{spikesMasud}
M.~S. Masud and R.~Borisyuk,
\newblock ``Statistical technique for analysing functional connectivity of
  multiple spike trains,''
\newblock {\em Journal of Neuroscience Methods}, vol. 196, no. 1, pp. 201--219,
  2011.

\bibitem{netrapalli2012learning}
Praneeth Netrapalli and Sujay Sanghavi,
\newblock ``Learning the graph of epidemic cascades,''
\newblock in {\em ACM SIGMETRICS Performance Evaluation Review}. ACM, 2012,
  vol.~40, pp. 211--222.

\bibitem{altarelli2014patient}
Fabrizio Altarelli, Alfredo Braunstein, Luca Dall'Asta, Alessandro Ingrosso,
  and Riccardo Zecchina,
\newblock ``The patient-zero problem with noisy observations,''
\newblock {\em Journal of Statistical Mechanics: Theory and Experiment}, vol.
  2014, no. 10, pp. P10016, 2014.

\bibitem{kempe2003maximizing}
David Kempe, Jon Kleinberg, and {\'E}va Tardos,
\newblock ``Maximizing the spread of influence through a social network,''
\newblock in {\em Proceedings of the ninth ACM SIGKDD international conference
  on Knowledge discovery and data mining}. ACM, 2003, pp. 137--146.

\bibitem{kuperman2001small}
M.~Kuperman and G.~Abramson,
\newblock ``Small world effect in an epidemiological model,''
\newblock {\em Physical Review Letters}, vol. 86, no. 13, pp. 2909, 2001.

\bibitem{johansson1996speed}
Per Johansson,
\newblock ``Speed limitation and motorway casualties: a time series count data
  regression approach,''
\newblock {\em Accident Analysis \& Prevention}, vol. 28, no. 1, pp. 73--87,
  1996.

\bibitem{matteson2011forecasting}
David~S Matteson, Mathew~W McLean, Dawn~B Woodard, and Shane~G Henderson,
\newblock ``Forecasting emergency medical service call arrival rates,''
\newblock {\em The Annals of Applied Statistics}, pp. 1379--1406, 2011.

\bibitem{rydberg1999modelling}
Tina~Hviid Rydberg and Neil Shephard,
\newblock ``A modelling framework for the prices and times of trades made on
  the new york stock exchange,''
\newblock Tech. {R}ep., Nuffield College, 1999,
\newblock Working Paper W99-14.

\bibitem{ait2010modeling}
Y.~A{\"\i}t-Sahalia, J.~Cacho-Diaz, and R.~J.~A. Laeven,
\newblock ``Modeling financial contagion using mutually exciting jump
  processes,''
\newblock Tech. {R}ep., National Bureau of Economic Research, 2010.

\bibitem{chavez2012high}
V.~Chavez-Demoulin and J.~A. McGill,
\newblock ``High-frequency financial data modeling using {H}awkes processes,''
\newblock {\em Journal of Banking \& Finance}, vol. 36, no. 12, pp. 3415--3426,
  2012.

\bibitem{cameron2013regression}
A~Colin Cameron and Pravin~K Trivedi,
\newblock {\em Regression analysis of count data}, vol.~53,
\newblock Cambridge university press, 2013.

\bibitem{raginsky_OCP}
M.~Raginsky, R.~Willett, C.~Horn, J.~Silva, and R.~Marcia,
\newblock ``Sequential anomaly detection in the presence of noise and limited
  feedback,''
\newblock {\em IEEE Transactions on Information Theory}, vol. 58, no. 8, pp.
  5544--5562, 2012.

\bibitem{silva:pami}
J.~Silva and R.~Willett,
\newblock ``Hypergraph-based anomaly detection in very large networks,''
\newblock {\em IEEE Transactions on Pattern Analysis and Machine Intelligence},
  vol. 31, no. 3, pp. 563--569, 2009,
\newblock {doi:10.1109/TPAMI.2008.232}.

\bibitem{BertozziHawkes}
A.~Stomakhin, M.~B. Short, and A.~Bertozzi,
\newblock ``Reconstruction of missing data in social networks based on temporal
  patterns of interactions,''
\newblock {\em Inverse Problems}, vol. 27, no. 11, 2011.

\bibitem{HellerHawkes}
C.~Blundell, K.~A. Heller, and J.~M. Beck,
\newblock ``Modelling reciprocating relationships with {H}awkes processes,''
\newblock in {\em Proc. NIPS}, 2012.

\bibitem{zhouZhaSongHawkes}
K.~Zhou, H.~Zha, and L.~Song,
\newblock ``Learning social infectivity in sparse low-rank networks using
  multi-dimensional {H}awkes processes,''
\newblock in {\em Proceedings of the 16th International Conference on
  Artificial Intelligence and Statistics (AISTATS)}, 2013.

\bibitem{huang2003short}
Shyh-Jier Huang and Kuang-Rong Shih,
\newblock ``Short-term load forecasting via arma model identification including
  non-{G}aussian process considerations,''
\newblock {\em Power Systems, IEEE Transactions on}, vol. 18, no. 2, pp.
  673--679, 2003.

\bibitem{hawkesEarthquake}
D.~Vere-Jones and T.~Ozaki,
\newblock ``Some examples of statistical estimation applied to earthquake
  data,''
\newblock {\em Ann. Inst. Statist. Math.}, vol. 34, pp. 189--207, 1982.

\bibitem{ogata1999seismicity}
Y.~Ogata,
\newblock ``Seismicity analysis through point-process modeling: A review,''
\newblock {\em Pure and Applied Geophysics}, vol. 155, no. 2-4, pp. 471--507,
  1999.

\bibitem{brannas1994time}
Kurt Br{\"a}nn{\"a}s and Per Johansson,
\newblock ``Time series count data regression,''
\newblock {\em Communications in Statistics-Theory and Methods}, vol. 23, no.
  10, pp. 2907--2925, 1994.

\bibitem{macdonald1997hidden}
Iain~L MacDonald and Walter Zucchini,
\newblock {\em Hidden Markov and other models for discrete-valued time series},
  vol. 110,
\newblock CRC Press, 1997.

\bibitem{zeger1988regression}
Scott~L Zeger,
\newblock ``A regression model for time series of counts,''
\newblock {\em Biometrika}, vol. 75, no. 4, pp. 621--629, 1988.

\bibitem{jorgensen1999state}
Bent J{\o}rgensen, Soren Lundbye-Christensen, PX-K Song, and Li~Sun,
\newblock ``A state space model for multivariate longitudinal count data,''
\newblock {\em Biometrika}, vol. 86, no. 1, pp. 169--181, 1999.

\bibitem{fahrmeir2013multivariate}
Ludwig Fahrmeir and Gerhard Tutz,
\newblock {\em Multivariate statistical modelling based on generalized linear
  models},
\newblock Springer Science \& Business Media, 2013.

\bibitem{grunwald2000theory}
Gary~K Grunwald, Rob~J Hyndman, Leanna Tedesco, and Richard~L Tweedie,
\newblock ``Theory \& methods: Non-{G}aussian conditional linear {AR} (1)
  models,''
\newblock {\em Australian \& New Zealand Journal of Statistics}, vol. 42, no.
  4, pp. 479--495, 2000.

\bibitem{benjamin2003generalized}
Michael~A Benjamin, Robert~A Rigby, and D~Mikis Stasinopoulos,
\newblock ``Generalized autoregressive moving average models,''
\newblock {\em Journal of the American Statistical association}, vol. 98, no.
  461, pp. 214--223, 2003.

\bibitem{gourieroux2005autoregressive}
Christian Gouri{\'e}roux and Joann Jasiak,
\newblock ``Autoregressive gamma processes,''
\newblock {\em Les Cahiers du CREF of HEC Montr{\'e}al Working Paper}, , no.
  05-03, 2005.

\bibitem{fokianos2009poisson}
Konstantinos Fokianos, Anders Rahbek, and Dag Tj{\o}stheim,
\newblock ``Poisson autoregression,''
\newblock {\em Journal of the American Statistical Association}, vol. 104, no.
  488, pp. 1430--1439, 2009.

\bibitem{zhu2011estimation}
Fukang Zhu and Dehui Wang,
\newblock ``Estimation and testing for a {P}oisson autoregressive model,''
\newblock {\em Metrika}, vol. 73, no. 2, pp. 211--230, 2011.

\bibitem{fokianos2011log}
Konstantinos Fokianos and Dag Tj{\o}stheim,
\newblock ``Log-linear {P}oisson autoregression,''
\newblock {\em Journal of Multivariate Analysis}, vol. 102, no. 3, pp.
  563--578, 2011.

\bibitem{hawkes1}
A.~G. Hawkes,
\newblock ``Point spectra of some self-exciting and mutually-exciting point
  processes,''
\newblock {\em Journal of the Royal Statistical Society. Series B
  (Methodological)}, vol. 58, pp. 83--90, 1971.

\bibitem{hawkes2}
A.~G. Hawkes,
\newblock ``Point spectra of some mutually-exciting point processes,''
\newblock {\em Journal of the Royal Statistical Society. Series B
  (Methodological)}, vol. 33, pp. 438--443, 1971.

\bibitem{PointProcesses}
D.~J. Daley and D.~Vere-Jones,
\newblock {\em An introduction to the theory of point processes, Vol. I:
  Probability and its Applications},
\newblock Springer-Verlag, New York, second edition, 2003.

\bibitem{hansenReynaud2015}
Niels~Richard Hansen, Patricia Reynaud-Bouret, and Vincent Rivoirard,
\newblock ``{LASSO} and probabilistic inequalities for multivariate point
  processes,''
\newblock {\em Bernoulli}, vol. 21, no. 1, pp. 83--143, 02 2015.

\bibitem{bacry2015}
Muzy Bacry, Gaiffas,
\newblock ``A generalization error bound for sparse and low-rank multivariate
  hawkes processes,''
\newblock arXiv:1501.00725, 2015.

\bibitem{heinen2003modelling}
Andr{\'e}as Heinen,
\newblock ``Modeling time series count data: an autoregressive conditional
  {P}oisson model,''
\newblock {\em Available at SSRN 1117187}, 2003.

\bibitem{zhu2011negative}
Fukang Zhu,
\newblock ``A negative binomial integer-valued garch model,''
\newblock {\em Journal of Time Series Analysis}, vol. 32, no. 1, pp. 54--67,
  2011.

\bibitem{zhu2012modeling}
Fukang Zhu,
\newblock ``Modeling time series of counts with {COM}-poisson {INGARCH}
  models,''
\newblock {\em Mathematical and Computer Modelling}, vol. 56, no. 9, pp.
  191--203, 2012.

\bibitem{zhu2012modeling_b}
Fukang Zhu,
\newblock ``Modeling overdispersed or underdispersed count data with
  generalized {P}oisson integer-valued garch models,''
\newblock {\em Journal of Mathematical Analysis and Applications}, vol. 389,
  no. 1, pp. 58--71, 2012.

\bibitem{willett:density}
R.~Willett and R.~Nowak,
\newblock ``Multiscale {P}oisson intensity and density estimation,''
\newblock {\em IEEE Transactions on Information Theory}, vol. 53, no. 9, pp.
  3171--3187, 2007,
\newblock {doi:10.1109/TIT.2007.903139}.

\bibitem{dense_pcs}
M.~Raginsky, R.~Willett, Z.~Harmany, and R.~Marcia,
\newblock ``Compressed sensing performance bounds under {P}oisson noise,''
\newblock {\em IEEE Transactions on Signal Processing}, vol. 58, no. 8, pp.
  3990--4002, 2010,
\newblock {arXiv:0910.5146}.

\bibitem{expander_PCS}
M.~Raginsky, S.~Jafarpour, Z.~Harmany, R.~Marcia, R.~Willett, and
  R.~Calderbank,
\newblock ``Performance bounds for expander-based compressed sensing in
  {P}oisson noise,''
\newblock {\em IEEE Transactions on Signal Processing}, vol. 59, no. 9, 2011,
\newblock {arXiv:1007.2377}.

\bibitem{JiangWillettRaskutti15}
X.~Jiang, R.~Willett, and G.~Raskutti,
\newblock ``Minimax rates of estimation for high-dimensional linear regression
  over $\ell_q$-balls,''
\newblock {\em IEEE Transactions on Information Theory}, vol. 61, pp.
  4458--4474, 2015.

\bibitem{geer08}
S.~van~de Geer,
\newblock ``High-dimensional generalized linear models and the {LASSO},''
\newblock {\em Annals of Statistics}, vol. 36, pp. 614--636, 2008.

\bibitem{KolYua08}
V.~Koltchinskii and M.~Yuan,
\newblock ``Sparse recovery in large ensembles of kernel machines,''
\newblock in {\em Proceedings of COLT}, 2008.

\bibitem{Meier09}
L.~Meier, S.~van~de Geer, and P.~Buhlmann,
\newblock ``High-dimensional additive modeling,''
\newblock {\em Annals of Statistics}, vol. 37, pp. 3779--3821, 2009.

\bibitem{Neg10}
S.~Negahban, P.~Ravikumar, M.~J. Wainwright, and B.~Yu,
\newblock ``A unified framework for high-dimensional analysis of {M}-estimators
  with decomposable regularizers,''
\newblock {\em Statistical Science}, vol. 27, no. 4, pp. 538--557, 2010.

\bibitem{RasWaiYu11}
G.~Raskutti, M.~J. Wainwright, and B.~Yu,
\newblock ``Minimax rates of estimation for high-dimensional linear regression
  over $\ell_q$-balls,''
\newblock {\em IEEE Transactions on Information Theory}, vol. 57, pp.
  6976--6994, 2011.

\bibitem{RasWaiYu12}
G.~Raskutti, M.~J. Wainwright, and B.~Yu,
\newblock ``Minimax-optimal rates for sparse additive models over kernel
  classes via convex programming,''
\newblock {\em Journal of Machine Learning Research}, vol. 13, pp. 398--427,
  2012.

\bibitem{Zhao06}
P.~Zhao and B.~Yu,
\newblock ``On model selection consistency of {LASSO},''
\newblock {\em Journal of Machine Learning Research}, vol. 7, pp. 2541--2567,
  2006.

\bibitem{BvdG2011}
P.~B\"uhlmann and S.~van~de Geer,
\newblock {\em Statistics for High-Dimensional Data: Methods, Theory and
  Applications},
\newblock Springer, 2011.

\bibitem{BasuMichail15}
S.~Basu and G.~Michailidis,
\newblock ``Regularized estimation in sparse high-dimensional time series
  models,''
\newblock {\em Annals of Statistics}, vol. 43, no. 4, pp. 1535--1567, 2015.

\bibitem{jiang2015data}
X.~Jiang, P.~Reynaud-Bouret, V.~Rivoirard, L.~Sansonnet, and R.~Willett,
\newblock ``A data-dependent weighted {LASSO} under {P}oisson noise,''
\newblock {\em arXiv preprint arXiv:1509.08892}, 2015.

\bibitem{BiRiTsy08}
P.~Bickel, Y.~Ritov, and A.~Tsybakov,
\newblock ``Simultaneous analysis of {L}asso and {D}antzig selector,''
\newblock {\em Annals of Statistics}, vol. 37, no. 4, pp. 1705--1732, 2009.

\bibitem{VDGMartingale}
S.~van~de Geer,
\newblock {\em Empirical Process Techniques for Dependent Data},
\newblock Springer-Verlag, New York, NY, 2002.

\bibitem{RakhlinTewari}
A.~Rakhlin, K.~Sridharan, and A.~Tewari,
\newblock ``Sequential complexities and uniform martingale laws of large
  numbers,''
\newblock {\em Probability Theory and Related Fields}, vol. 1, no. 161, pp.
  111--153, February 2015.

\bibitem{BarAlb99}
Albert-L{\'a}szl{\'o} Barab{\'a}si and R{\'e}ka Albert,
\newblock ``Emergence of scaling in random networks,''
\newblock {\em Science}, vol. 286, no. 5439, pp. 509--512, 1999.

\bibitem{HoudreReynaudBouret03}
Christian Houdr{\'e} and Patricia Reynaud-Bouret,
\newblock ``Exponential inequalities, with constants, for {U}-statistics of
  order two,''
\newblock in {\em Stochastic inequalities and applications}, pp. 55--69.
  Springer, 2003.

\bibitem{BobkovLedoux}
S.~G. Bobkov and M.~Ledoux,
\newblock ``On modified logarithmic {S}oboloev inequalities for {B}ernoulli and
  {P}oisson measures,''
\newblock {\em Journal of Functional Analysis}, vol. 156, pp. 347--365, 1998.

\bibitem{LevinPeresBook}
D.~A. Levin, Y.~Peres, and E.~L. Wilmer,
\newblock {\em Markov Chains and Mixing Times},
\newblock American Mathematical Society, 2008.

\bibitem{Pollard84}
D.~Pollard,
\newblock {\em Convergence of {S}tochastic {P}rocesses},
\newblock Springer-Verlag, New York, 1984.

\end{thebibliography}
\end{document}